\documentclass{article}
\usepackage{natbib}

\usepackage{arxiv}

\usepackage[T1]{fontenc}    
\usepackage{hyperref}       
\usepackage{url}            
\usepackage{booktabs}       
\usepackage{amsfonts}       
\usepackage{nicefrac}       
\usepackage{microtype}      
\usepackage{lipsum}		
\usepackage{graphicx}
\usepackage{natbib}
\usepackage{doi}
\usepackage{amssymb} 
\usepackage{multirow}
\usepackage{array}
\usepackage{booktabs}

\usepackage{amsmath}   
\usepackage{amssymb}   
\usepackage{booktabs}  

\usepackage{tabularx}    
\usepackage{booktabs}    
\usepackage{makecell}    

\usepackage{multirow}    

\usepackage{graphicx}    
\usepackage{caption}
\usepackage{subcaption}  
\usepackage{geometry}
\usepackage[toc,page]{appendix}  

\usepackage{natbib}  
\newcommand{\revision}[1]{\textbf{#1}}

\title{\textbf{VTouch\textsuperscript{++}}: A Multimodal Dataset with Vision-Based Tactile Enhancement for Bimanual Manipulation}

\author{%
  Qianxi Hua$^{*,1}$,\;
  Xinyue Li$^{*,1}$,\;
  Zheng Yan $^{*,1,2}$,\;
  Yang Li$^{1}$,\;
  Chi Zhang$^{1,3}$,\;
  Yongyao Li$^{\dagger,1}$,\;
  Yufei Liu$^{\dagger,1}$ \\[0.6em]
  {\small $^1$Humanoid Robot (Shanghai) Co., Ltd. , Shanghai, China} \\
  {\small $^2$Shanghai Research Institute for Intelligent Autonomous Systems, Tongji University, Shanghai, China} \\
  {\small $^3$School of Astronautics, Harbin Institute of Technology, Harbin, China} \\[0.4em]
  {\small $^*$Equal contribution\quad$^\dagger$Corresponding author} \\
  {\small Yongyao Li: \texttt{yongyao.li@openloong.net}\quad Yufei Liu: \texttt{liuyufei@openloong.net}}
}

\renewcommand{\shorttitle}{\textit{arXiv} Template}

\hypersetup{
pdftitle={VTouch++: A Multimodal Dataset with Vision-Based Tactile Enhancement for Bimanual Manipulation},
pdfsubject={Robotics, Embodied AI, Multimodal Learning, Visuotactile Perception},
pdfauthor={Qianxi Hua, Xinyue Li, Zheng Yan, Yang Li, Chi Zhang, Yongyao Li, Yufei Liu},
pdfkeywords={visuotactile learning, multimodal dataset, bimanual manipulation, tactile sensing, cross-modal retrieval, policy learning},
}

\begin{document}
\maketitle

\begin{figure}[htbp]
    \centering
    \includegraphics[width=0.9\textwidth]{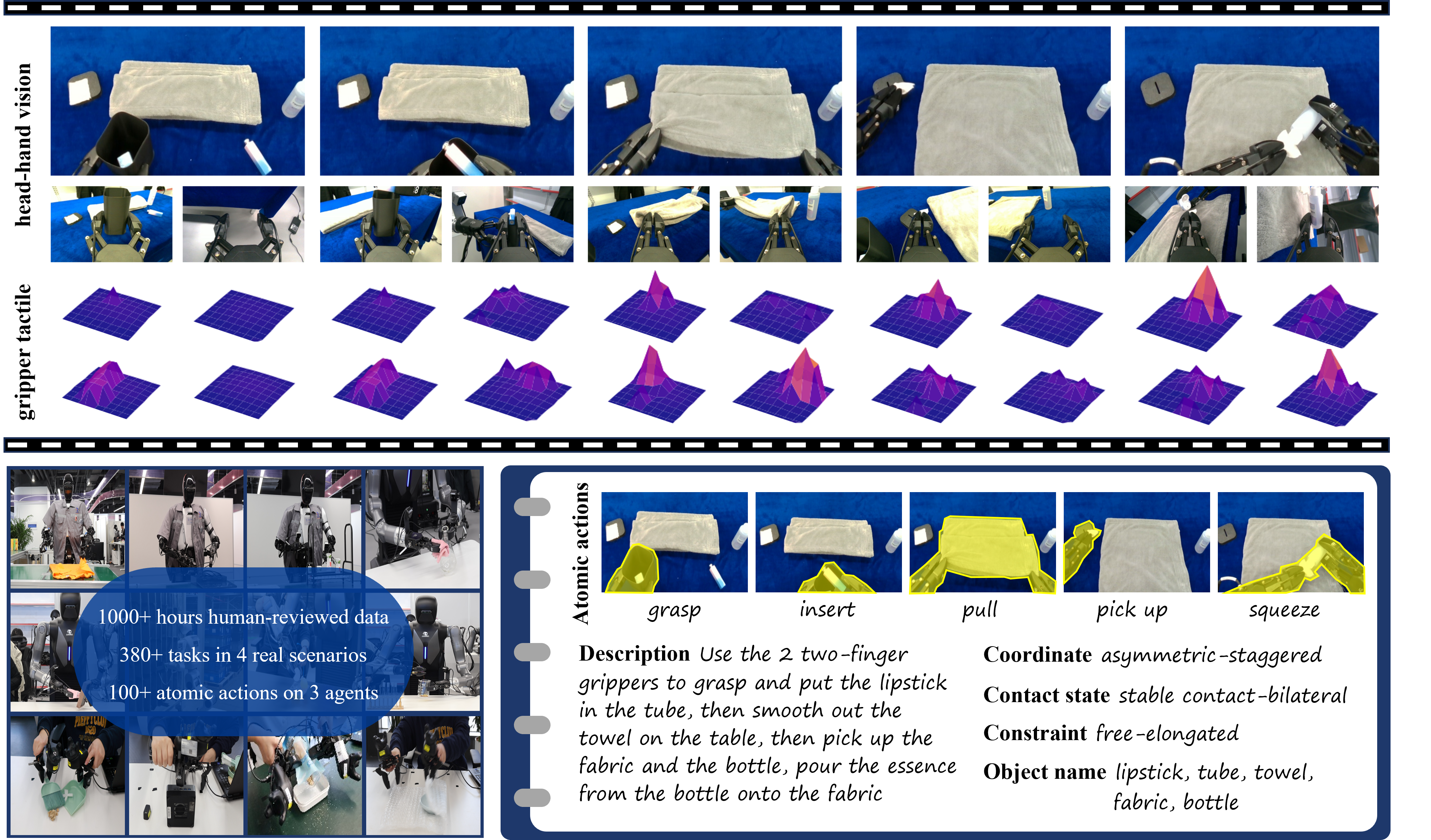}
    \caption{\textbf{Dataset Overview.} The proposed multimodal bimanual manipulation dataset captures synchronized proprioception, multi-view RGB-D observations, and high-resolution fingertip tactile signals from multiple robot embodiments. The dataset comprises over 380+ bimanual tasks with 100+ atomic action compositions, providing a foundational resource for contact-intensive manipulation research.}
    \label{fig:dataset_overview}
\end{figure}

\begin{abstract}
Embodied intelligence has advanced rapidly in recent years; however, bimanual manipulation-especially in contact-rich tasks—remains challenging. This is largely due to the lack of datasets with rich physical interaction signals, systematic task organization, and sufficient scale. To address these limitations, we introduce the VTOUCH dataset. It leverages vision based tactile sensing to provide high-fidelity physical interaction signals, adopts a matrix-style task design to enable systematic learning, and employs automated data collection pipelines covering real-world, demand-driven scenarios to ensure scalability. To further validate the effectiveness of the dataset, we conduct extensive quantitative experiments on cross-modal retrieval as well as real-robot evaluation. Finally, we demonstrate real-world performance through generalizable inference across multiple robots, policies, and tasks.
\end{abstract}


\section{Introduction}
Bimanual manipulation is a core capability for robots operating in human-centric environments such as household service, retail assistance, and industrial assembly. Compared to single-arm manipulation, bimanual tasks not only involve richer coordination patterns and stronger physical constraints, but also critically rely on multimodal perception—particularly the joint reasoning over visual and tactile signals to handle frequent and uncertain contact interactions. This places substantially higher demands on perception, control, and representation learning. While recent imitation learning and diffusion-based policies have demonstrated strong performance in vision-driven manipulation, their progress in contact-intensive bimanual settings remains fundamentally limited by the lack of large-scale datasets that are grounded in real-world physical interaction and jointly capture visual and tactile feedback.

Existing datasets exhibit complementary yet critical limitations. Human bimanual interaction datasets offer scale and behavioral diversity, but lack access to robot-specific proprioception, contact forces, and embodiment constraints. Synthetic or simulation-based robotic datasets provide precise state and force annotations, yet often suffer from sim-to-real discrepancies that hinder real-world deployment. Real-robot manipulation datasets, while physically realistic, are typically limited in scale, sensing modalities, or task diversity particularly in the bimanual setting. Crucially, there is currently no large-scale, multimodal dataset for bimanual robot manipulation that jointly captures real-world robot proprioception, visual observations, and tactile sensing, while being collected from physically verified interactions.

To address this gap, we introduce a large-scale multimodal dataset constructed from real-world bimanual manipulation demonstrations, collected across heterogeneous platforms, including \textbf{bipedal humanoid robots} such as Qingloong, \textbf{wheeled humanoid robots} such as Wheelloong M1, and UMI-style mobile manipulators . The dataset synchronously records joint-level proprioception, multi-view RGB-D observations, and explicit fingertip tactile signals obtained from \cite{Li_2013} tactile sensors, enabling high-fidelity capture of contact-rich manipulation dynamics. By grounding all data in real hardware execution, the dataset avoids sim-to-real artifacts and provides a reliable foundation for learning and evaluation.

Beyond scale and sensing richness, our dataset is guided by a skill-axis design philosophy. Rather than organizing data around a fixed set of discrete task labels, we structure demonstrations along fundamental axes including bimanual coordination patterns, atomic manipulation actions, sensory modalities, and temporal organization. Over 300 bimanual tasks are represented as compositions of atomic actions under diverse coordination and contact conditions, enabling systematic recomposition and analysis without requiring ambiguous sub-trajectory segmentation.

Finally, we position this dataset as a bridge between theoretical modeling, real-robot data acquisition, and reproducible policy learning. We benchmark representative learning-based methods, including Action Chunking Transformers and diffusion-based policies with visual–tactile fusion, demonstrating the necessity of multimodal perception and explicit bimanual coordination for contact-intensive manipulation. Together, this dataset and its benchmarks aim to accelerate research toward robust and generalizable bimanual manipulation in real-world environments.

\section{Related Work}
\label{sec:headings}

We have surveyed existing datasets and methods across two domains to situate our contribution: datasets of physical interactions, particularly those involving bimanual coordination, and datasets for multimodal perception, particularly those involving tactile. An extended discussion and comparison are provided in \revision{Table 1}.

\subsection{Physical Interaction Datasets}

Human Bimanual Interaction Datasets. 

Large-scale first-person video datasets, such as Ego4D \cite{grauman2022ego4dworld3000hours} and Epic-Kitchens \cite{damen2020epickitchensdatasetcollectionchallenges}, capture a wide range of human activity scenarios, including household, outdoor, workplace, leisure, and more, while also featuring bimanual interactions, providing extensive task coverage and contextual understanding. However, they only offer RGB video data, lacking precise 3D pose annotations and any physical contact information, which makes them primarily suitable for high-level task understanding rather than learning contact-intensive control policies.

Optical motion capture systems address the accuracy issue, high-precision motion capture datasets. . Datasets such as GRAB \cite{Taheri_2020} capture whole-body grasping motions, containing full 3D shape and pose sequences of 10 subjects. While ARCTIC \cite{fan2024hold} extends this to complex bimanual object manipulation with synchronized RGB-D data streams. These works provide high-fidelity kinematic trajectories for both hands and objects. However, a key limitation is that they only record geometric motions without capturing the contact forces that cause those motions. The absence of such physical interaction signals makes direct transfer to robotics—where force modulation is critical—challenging.

Robotic Manipulation Datasets. 

Physical simulation enables the generation of large-scale datasets with perfect ground truth. DexGraspNet\cite{wang2023dexgraspnetlargescaleroboticdexterous} and its subsequent work DexGraspNet 2.0\cite{zhang2024dexgraspnet20learninggenerative} generate stable grasps via synthesis method which concludes a two-stage grasping and diffusion model approach. For bimanual coordination, BiDexHands\cite{chen2022humanlevelbimanualdexterousmanipulation} provides a high-performance simulation environment and policies. While their scale is virtually unlimited, these datasets are affected by the sim-to-real gap—discrepancies in dynamics, sensing, and rendering limit the transfer of policies to physical systems.

Data collected directly on target hardware, whether through teleoperation \cite{ze2025twist2} or autonomous exploration \cite{contributors2024agibotworldrepo}, provides real kinematics. However, scaling such collection is prohibitively expensive and slow, especially for high degree-of-freedom dual-arm systems. Furthermore, most real-world robot datasets primarily rely on visual observations and rarely integrate high-bandwidth tactile sensing. This lack of rich, synchronized proprioceptive-tactile-visual data hinders the advancement of learning granular, contact-aware manipulation skills. RoboNet\cite{DBLP:journals/corr/abs-1910-11215} presents an open large-scale multi-robot manipulation dataset. The dataset is gathered through autonomous random exploration and supports cross-robot and cross-scene pre-training and fine-tuning experiments, yet it still lacks tactile fusion, bimanual coordination, and complex task sequences.To overcome the cost and efficiency bottlenecks of real-robot collection, recent robot-free paradigms have emerged. For instance, FreeTacMan\cite{wu2025freetacmanrobotfreevisuotactiledata} designs a wearable visuo-tactile gripper operated directly by humans, combined with high-precision motion capture to record end-effector poses, thereby efficiently collecting large-scale, multimodal manipulation data. Such methods significantly improve collection efficiency and user experience, but the learned policies still need to be transferred to real robots, and their operational morphology (parallel gripper) differs from complex bimanual dexterous hands.

\subsection{Multimodal Perception datasets}

Contact sensing. Another line of work attempts to infer contact. ContactDB\cite{brahmbhatt2019contactdbanalyzingpredictinggrasp} and ContactPose\cite{brahmbhatt2020contactposedatasetgraspsobject} employ thermal imaging to map hand-object contact areas. While introducing a physical dimension, thermal imaging only provides binary contact masks without information about force magnitude and direction, and is susceptible to environmental thermal noise. Recent diffusion-based methods\cite{Christen_2024} can generate diverse hand-object interactions but lack a physical foundation.

Visual-Tactile Fusion. Tactile sensors, such as GelSight or DIGIT\cite{lambeta2024digit360}, provide high-resolution contact geometry and force cues. Some works have collected specialized tactile datasets\cite{fenganytouch}\cite{Li_2025}, though these are typically limited in scale and task-specific. A key challenge lies in the synchronized integration of tactile signals with vision and proprioception into large-scale, general-purpose manipulation datasets. Furthermore, addressing the heterogeneity of tactile sensor data, TacQuad \cite{feng2025learning} provides aligned contact data from four sensor types, establishing a basis for cross-sensor unified tactile representation. However, such perception datasets typically lack integration with continuous manipulation tasks and physical dynamics. Going a step further, the NeuralFeels\cite{suresh2024neuralfeels} system implements an online visuo-tactile SLAM framework, which reconstructs the shape and pose of unknown objects in real-time via neural fields, significantly improving tracking robustness under heavy occlusion. It serves as a comprehensive technical exemplar for multimodal perception and physical interaction modeling. However, its accompanying FeelSight dataset is limited in scale and primarily focuses on simple in-hand rotation tasks for a single hand, lacking coverage of bimanual coordinated manipulation and more complex, long--horizon task sequences. Meanwhile, V-HOP \cite{Li_2025} introduces a learning-based visuo-haptic pose tracking framework. By leveraging a unified point cloud representation and a Transformer-based fusion mechanism, it achieves strong generalization capabilities across novel grippers, sensor types, and objects.

Physically Consistent Observation Reconstruction. An alternative to direct measurement involves inferring physical interactions through consistency with known physical laws. Recent approaches leverage physical simulation to convert purely kinematic demonstrations into physically plausible trajectories annotated with forces. DexCanvas\cite{xu2025dexcanvasbridginghumandemonstrations} pioneered this method for single-handed manipulation: it employs reinforcement learning to train a simulated hand to track captured object motion, while the physics simulator provides the resulting contact forces as ground truth. This reframes the ill-posed problem of "estimating forces from observations" into the well-posed problem of "controlling under physical constraints to replicate observed motion," thereby generating physically consistent annotations. However, this powerful paradigm has yet to be applied to the domain of bimanual robotic manipulation with real-world tactile sensing.

As summarized in Table 1, existing datasets either lack physical force annotations (human datasets), are affected by the sim-to-real gap (synthetic robotic datasets), or are limited in scale and modality (real-world robotic datasets). Critically, there is currently no large-scale, multimodal dataset for bimanual robotic manipulation that integrates real-world robot proprioception, vision, and tactile sensing with physically validated interaction annotations. Our work aims to fill this gap. We introduce a dataset constructed from real-world bimanual robot demonstrations, which synchronously records joint states, multi-view RGB-D, and fingertip tactile sensing.

\begin{table}[htbp]
\centering
\caption{Dataset Comparision Table}
\label{tab:lit_review}
\small
\begin{tabularx}{\textwidth}{>{\centering\arraybackslash}m{2.1cm} 
			>{\centering\arraybackslash}m{0.9cm} 
			>{\centering\arraybackslash}m{2.1cm} 
			>{\centering\arraybackslash}m{1.5cm} 
			>{\centering\arraybackslash}m{1.8cm} 
			>{\centering\arraybackslash}m{2.5cm} 
			>{\centering\arraybackslash}m{2.5cm}}
\toprule
\thead{Dataset} & \thead{Dual-\\arm} & \thead{Tactile\\Modality} & \thead{Physical\\Verification} & \thead{Scale} & \thead{Main\\Modality} & \thead{Annotation\\Source} \\
\midrule
\multicolumn{7}{c}{\textbf{Human Bimanual Interaction Datasets}} \\
\midrule
EGO4D & \checkmark & $\times$ & $\times$ & 3700+ hours & Video (1st-person + 3rd-person) & None \\
Epic-Kitchens & \checkmark & $\times$ & $\times$ & 100+ hours & RGB (1st-person) & None \\
GRAB & \checkmark & Binary & $\times$ & \makecell{10 participants\\4 action intents} & Motion Capture & Thermal imaging \\
ARCTIC & \checkmark & $\times$ & $\times$ & 2.16 hours & Motion Capture + RGB-D & Markers \\
\midrule
\multicolumn{7}{c}{\textbf{Robotic Manipulation Datasets}} \\
\midrule
BiDexHands & \checkmark & $\times$ & \checkmark (Sim) & 40K+ FPS (simulation) & State & Simulation \\
DexGraspNet 2.0 & $\times$ & $\times$ & \checkmark (Sim) & 427 million grasps & State & Optimization \\
AgiBot-World  & \checkmark & $\times$ & \checkmark & 2976.4 hours & \makecell{RGB-D/Teleop\\ + joints} & Robot \\
RoboNet & $\times$ & $\times$ & $\times$ & 15 million frames & RGB & Robot \\
FreeTacMan & $\times$ & Vision-based tactile & \checkmark & 3000k image pairs & Vision-based tactile & Vision-based tactile images \\

\midrule
\multicolumn{7}{c}{\textbf{Multimodal Perception Datasets}} \\
\midrule
DexCanvas & $\times$ & Binary & \checkmark (Sim) & 70+ hours (real) & Motion Capture & Markers \\
ContactDB & $\times$ & Binary (thermal) & $\times$ & 3.5 hours & RGB-D + thermal & Thermal imaging \\
ContactPose & $\times$ & Binary (thermal) & $\times$ & 2.9M frames & RGB-D + thermal & Thermal imaging \\
TacQuad & $\times$ & Vision-based tactile & -- & 72,606 frames & RGB + video & Vision-based tactile images \\
Feelsight & $\times$ & Vision-based tactile & \checkmark & About 35 min & \makecell{RGB-D + joints\\ + object pose} & \makecell{Vision-based tactile\\ image markers} \\
V-HOP & $\times$ & \makecell{Binary/\\Vision-based tactile} & $\times$ & About 1.55M images & RGB-D + tactile & \makecell{Binary/\\Vision-based \\tactile images} \\
\midrule
\textbf{Our Dataset} & \textbf{\checkmark} & \textbf{\makecell{High-res (real)\\ + simulation}} & \textbf{\checkmark} & \textbf{Large-scale} & \textbf{\makecell{RGB-D + joints\\ + tactile}} & \textbf{Robot + simulation} \\
\bottomrule
\end{tabularx}
\end{table}

\section{Dataset Construction}
\label{sec:others}

\subsection{Data Acquisition System Design}
Hardware Configuration: Multi-configuration physical dual-arm robot platform, visuotactile sensors, multi-view RGB-D cameras, synchronization trigger module.

Software Architecture: ROS 2 data stream synchronization, timestamp alignment, data storage format.

Human-Robot Interface: Supports teleoperation, motion recording, and motion playback.

The data acquisition system consists of three parts: a hardware platform, a software architecture, and a human‑machine interaction interface.

\subsubsection{Hardware Configuration}
The system is built upon a multi‑configuration physical dual‑arm robot platform, covering fixed dual‑arm systems, wheeled‑arm systems, and UMI (hand‑held grippers). All robots are connected via a unified hardware abstraction interface, which semantically aligns different hardware configurations at the state, action, and sensing levels. This enables data from multi‑configuration dual‑arm robots to be uniformly represented, processed, and learned. The system integrates head‑mounted and wrist‑mounted RGB‑D cameras along with fingertip tactile sensors. The latter are installed modularly on the end‑effectors and aligned with the robot coordinate system through a calibration procedure. To achieve high‑precision cross‑modal synchronization, robot proprioceptive data, cameras, and visual‑tactile sensors are acquired synchronously by a unified hardware triggering module.

\subsubsection{Software Architecture}
The system constructs a multimodal data‑flow pipeline based on ROS2. By combining hardware triggering and timestamp mechanisms, temporal alignment of visual, tactile, and proprioceptive data is achieved. For data streams with different sampling rates, interpolation and alignment strategies are employed for uniform packaging. Consistency verification and anomaly detection mechanisms are introduced during both acquisition and post‑processing stages to ensure data integrity and stability.

\subsubsection{Human‑Machine Interaction Interface}
The system supports teleoperation teaching, action recording, and playback on real robots. During teaching, real‑time visual and state feedback is provided. After acquisition, the demonstrator can review and filter trajectories through playback verification, thereby ensuring the quality of demonstrations. Meanwhile, all recorded demonstration trajectories can be accurately reproduced on physical robot platforms, providing a unified execution basis for subsequent policy learning and benchmark evaluation.

\begin{figure}[htbp]
    \centering
    \includegraphics[width=0.85\textwidth]{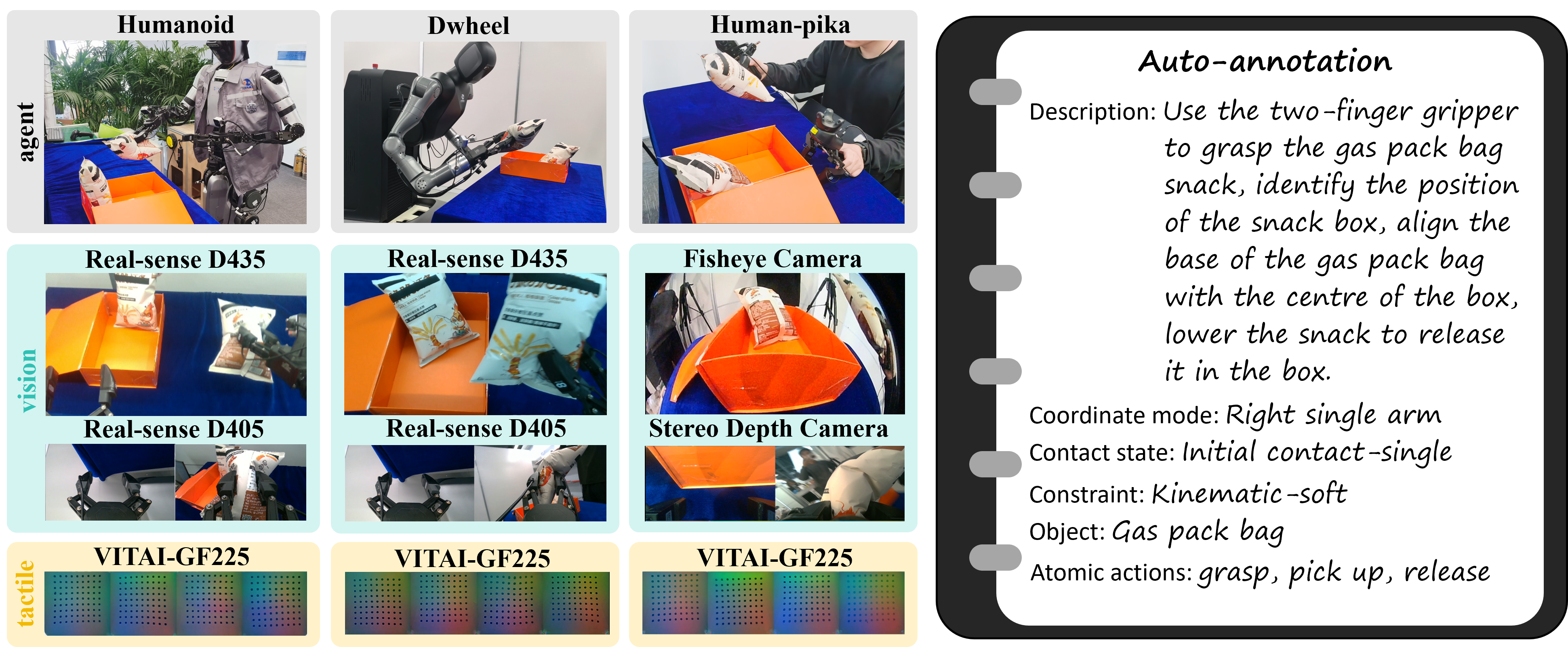}
    \caption{\textbf{Cross-Embodiment Data Collection.} The data acquisition system supports multiple robot embodiments including fixed dual-arm platforms, wheeled-arm systems, and UMI-style mobile manipulators. All platforms are connected via a unified hardware abstraction interface that semantically aligns different hardware configurations at the state, action, and sensing levels.}
    \label{fig:cross_embodiment}
\end{figure}

\subsection{Task Design}

\begin{figure}[htbp]
    \centering
    \includegraphics[width=0.85\textwidth]{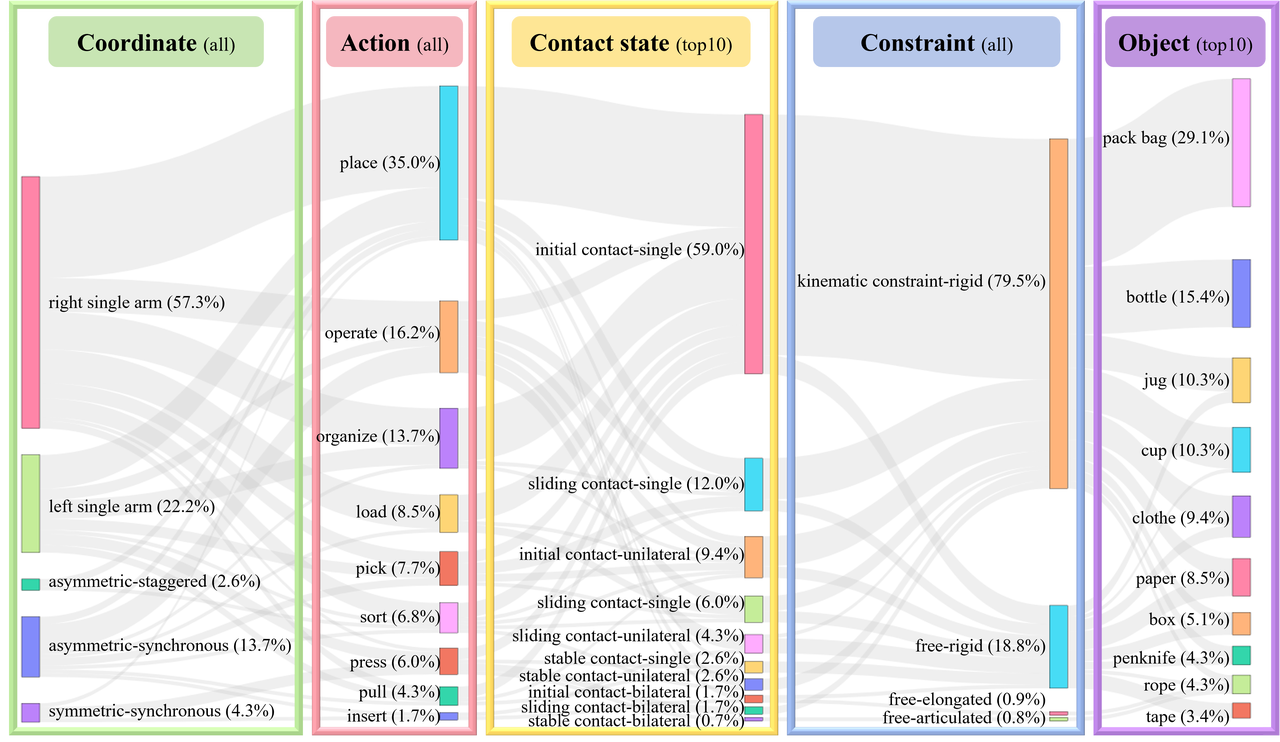}
    \caption{\textbf{Task Classification Framework.} The skill-axis framework categorizes bimanual manipulation tasks along six orthogonal dimensions: bimanual coordination structure, atomic action types, contact and tactile modes, object and geometry properties, perception modality requirements, and task composition hierarchy.}
    \label{fig:task_classification}
\end{figure}

\begin{table}[htbp]
\centering
\caption{Skill Axes Classification Framework}
\label{tab:skill_axes}
\begin{tabularx}{\textwidth}{>{\centering\arraybackslash}m{2.2cm}>{\raggedright\arraybackslash}m{3.2cm}>{\raggedright\arraybackslash}m{4.2cm}>{\raggedright\arraybackslash}X}
\toprule
\textbf{Axis Category} & \textbf{Skill Axis} & \textbf{Discrete Values / Examples} & \textbf{Design Motivation} \\
\midrule

\multirow{2}{*}{\centering\makecell{A. Bimanual\\Coordination\\Structure}} 
& Coordination Pattern & Symmetric / Asymmetric / Master–Slave / Sequential 
& To clarify bimanual information non-reducibility (non-reducible to single-arm) \\
\cmidrule(lr){2-4}
& Temporal Coupling & Synchronous / Staggered / Alternating 
& To capture temporal structure differences without relying on sub-trajectories \\
\midrule

\centering B. Atomic Action Types 
& Atomic Action Type & grasp, hold, pull, push, rotate, align, insert, release, etc. 
& As minimal interpretable units, replacing sub-trajectories \\
\midrule

\multirow{3}{*}{\centering\makecell{C. Contact and\\Tactile}} 
& Fingertip Contact State & none / initial contact / stable contact / sliding contact 
& To explicitly model fingertip tactile phases \\
\cmidrule(lr){2-4}
& Force Regulation Mode & maintain / increase / modulate / compliant 
& To rely solely on fingertip tactile feedback without force sensors \\
\cmidrule(lr){2-4}
& Contact Asymmetry & unilateral / bilateral 
& To represent tactile information asymmetry between hands \\
\midrule

\multirow{3}{*}{\centering\makecell{D. Object and\\Geometry}} 
& Object Count & single object / two-object interaction 
& To support cooperative assembly and alignment \\
\cmidrule(lr){2-4}
& Constraint Type & free / kinematic constraint / insertion-like 
& To reflect task difficulty hierarchy \\
\cmidrule(lr){2-4}
& Object Geometry & rigid / elongated / articulated 
& Not requiring fine classification but affecting policy design \\
\midrule

\multirow{2}{*}{\centering\makecell{E. Perception\\Modality}} 
& Visual Availability & full-view / partial-occlusion 
& To support multimodal necessity analysis \\
\cmidrule(lr){2-4}
& Tactile Dependency & optional / critical 
& To explain failure scenarios of vision-only approaches \\
\midrule

\centering\makecell{F. Task\\Composition\\Level} 
& Atomic Task Count & 1 / 2–3 / long-horizon (>3) 
& To use "atomic task concatenation" instead of sub-trajectories \\
\bottomrule
\end{tabularx}
\end{table}

Task Categories:Catering Service, Household and Furniture Care, Commercial and Pharmaceutical Scenarios, Industrial Manufacturing.

Object Set: Covers daily objects, tools, industrial parts, etc.

Scene Diversity: Variations in lighting, occlusion, and object poses.

To systematically construct a multi‑configuration dual‑arm multimodal manipulation task suite, this paper proposes a task generation framework based on the combination of multidimensional skill axes. Within this framework, each task is defined as a “minimally executable instance,” which is uniquely characterized by selecting specific values along multiple orthogonal skill axes, thereby capturing the task’s structural and semantic properties.

The framework is organized around three core axes that fundamentally describe the cooperative action, atomic operations, and contact interaction:

A. Bimanual Coordination Structure: This includes the Coordination Pattern (e.g., Symmetric, Asymmetric, Master–Slave, Sequential) and Temporal Coupling (e.g., Synchronous, Staggered, Alternating). It is designed to explicitly model the structural and temporal interdependencies that are irreducible to single-arm actions.

B. Atomic Action Type: Tasks are decomposed into sequences of minimal, interpretable operational units—such as grasp, hold, pull, push, rotate, align, insert, release—replacing traditional sub-trajectory descriptions to enhance semantic clarity and transferability.

C. Contact \& Tactile Mode: This axis encompasses the Fingertip Contact State (e.g., none, initial contact, stable contact, sliding contact), Force Regulation Mode, and Contact Asymmetry (unilateral/bilateral). It explicitly models the interactive process at the tactile level, supporting the development of closed-loop policies based primarily on tactile feedback.

Supplementary axes are introduced to ensure comprehensive task description and enhance the framework's extensibility:

D. Object \& Geometry: Includes Object Count, Constraint Type (e.g., free motion, kinematic constraint, insertion-like), and Object Geometry category, reflecting the physical and geometric complexity of the task.

E. Perception Modality: Evaluates Visual Availability (e.g., full-view, partial-occlusion) and Tactile Dependency (optional/critical), facilitating analysis of the necessity for multimodal perception across different tasks.

F. Task Composition Hierarchy: Characterizes task horizon and compositional complexity via the Atomic Task Count, supporting systematic coverage from single atomic tasks to long-horizon combinations.

This framework enables the generation of diverse, well-structured task instances through flexible axis combinations. For example, the task "collaboratively aligning and screwing a cap onto a bottle" can be described as:

A: Coordination Pattern = Master–Slave, Temporal Coupling = Synchronous.

B: Atomic Action Sequence = grasp → hold → rotate.

C: Fingertip Contact State = stable contact + sliding, Contact Asymmetry = bilateral.

By providing this structured semantic foundation, the framework supports the systematic generation, categorization, and analysis of dual-arm manipulation tasks, paving the way for subsequent skill learning, transfer, and benchmarking.

\subsection{Data preprocessing and quality control}
\subsubsection{Data review and initial labeling}
Prior to model training, we perform dataset auditing and weak annotation to identify potential anomalies and sensor artifacts. Specifically, sliding-window statistics are computed on each sensor channel to detect distribution shifts and abrupt deviations, enabling the screening of abnormal signal patterns and inconsistent interactions. Based on these detections, heuristic rules informed by physical constraints are applied to generate event-level weak labels, such as contact onset, anomalous interactions, or failed demonstrations.
This procedure follows common practices in time-series anomaly detection and robotic sensor monitoring, where weakly supervised methods are widely adopted to ensure data quality in the absence of large-scale manual annotations.
Prior to model training, we perform dataset auditing and weak annotation to identify potential anomalies and sensor artifacts. Specifically, sliding-window statistics are computed on each sensor channel to detect distribution shifts and abrupt deviations, enabling the screening of abnormal signal patterns and inconsistent interactions. Based on these detections, heuristic rules informed by physical constraints are applied to generate event-level weak labels, such as contact onset, anomalous interactions, or failed demonstrations.
This procedure follows common practices in time-series anomaly detection and robotic sensor monitoring, where weakly supervised methods are widely adopted to ensure data quality in the absence of large-scale manual annotations.

Additionally, temporal consistency across multiple modalities is leveraged as an auxiliary criterion to cross-validate anomalous cases that are difficult to identify using a single modality, thereby further improving the reliability of weak annotations.

\subsubsection{Automatic anomaly detection}
We adopt a channel-wise statistical anomaly detection approach\cite{w13131862} based on sliding window mean and variance estimation, where samples exceeding an nnn-sigma threshold are flagged as anomalies. This method belongs to the class of parametric statistical time-series anomaly detection techniques, which are widely used as interpretable and efficient baselines in sensor-based and robotic monitoring systems. Similar statistical thresholding approaches have been extensively applied to force/torque and tactile signal monitoring for collision detection, contact state change detection, and fault diagnosis in robotic manipulation.

\subsubsection{Temporal Alignment and Resampling}
Due to heterogeneous sampling rates and communication delays across visual, tactile, proprioceptive, and control streams, raw sensor data are not temporally aligned. Such misalignment can degrade downstream multimodal representation learning and policy training.
We use the robot control loop as the reference timeline and align all modalities based on their timestamps provided by ROS2. All streams are resampled to a unified frequency.

Visual observations are aligned at the frame level, while proprioceptive and tactile signals are resampled using linear interpolation or zero-order hold, depending on their physical semantics. Control commands are resampled using zero-order hold to preserve piecewise-constant actuation.

After temporal alignment, demonstrations with excessive missing data or severe temporal jitter are filtered out. Aligned trajectories are further segmented into task-relevant episodes for downstream learning.
\subsubsection{Demonstration Filtering and Segmentation}

\subsection{Dataset Construction}

Our dataset is collected from real-world bimanual manipulation tasks on the OpenLoong platform, featuring multi-modal sensory inputs including RGB cameras, visual-tactile sensors, and robot proprioception. This section describes the data collection and processing pipeline.

\subsubsection{Data Collection Platform}

The OpenLoong bimanual manipulation platform features:

\begin{itemize}
    \item \textbf{Dual 7-DOF Arms}: Two collaborative robot arms with 14 joints total
    \item \textbf{Three RGB-D Cameras}: Left, right, and head viewpoints for spatial awareness
    \item \textbf{Four Visual-Tactile Sensors}: GelSight-style sensors on both end effectors for contact perception
    \item \textbf{State Feedback}: Joint positions, velocities, end-effector poses, and gripper states
\end{itemize}

\subsubsection{Observation Modalities}

The dataset records observations from multiple sensory modalities:

\begin{table}[h]
\centering
\begin{tabular}{>{\centering\arraybackslash}m{2.5cm}
                >{\centering\arraybackslash}m{3cm}
                >{\centering\arraybackslash}m{5cm}}
\hline
Modality & Keys & Dimension \\
\hline
RGB Camera & camera\_left, camera\_right, head\_camera & $3 \times H \times W$ \\
Visual-Tactile & \makecell{tactile\_left \\ tactile\_right} & $3 \times H \times W$ \\
Joint State & - & 14 (positions) + 14 (velocities) \\
End-Effector & - & $7 \times 2$ (pose per arm) \\
Gripper & - & 2 (width per arm) \\
\hline
\end{tabular}
\end{table}

\subsubsection{Action Space}

The action space corresponds to the bimanual configuration:

\begin{itemize}
    \item \textbf{Joint Control}: 14-dimensional joint position commands
    \item \textbf{End-Effector Control}: 7-DOF pose per arm (position + quaternion)
    \item \textbf{Gripper Control}: Binary open/close per arm
\end{itemize}

Action can be specified as absolute positions or relative deltas.

\subsubsection{Data Processing Pipeline}

Raw sensor data is processed to create training-ready observations:

\begin{enumerate}
    \item \textbf{Temporal Alignment}: Synchronize all sensors to 30Hz sampling rate
    \item \textbf{Frame Stacking}: Stack $n_{\text{obs\_steps}}$ consecutive frames for temporal context
    \item \textbf{Normalization}: Apply per-modality normalization (mean-std or min-max)
    \item \textbf{Quality Filter}: Remove episodes with missing data or artifacts
\end{enumerate}

\subsubsection{Dataset Statistics}

\begin{table}[h]
\centering
\begin{tabular}{lc}
\hline
Metric & Value \\
\hline
Total Episodes & $\sim$120,000+ \\
Trajectory Duration & 10-60s per episode \\
Sampling Rate & 30Hz \\
Total Frames & $\sim$36M+ \\
\end{tabular}
\end{table}

The dataset is stored in RoboMimic or LeRobot format (video-based) with metadata for efficient training.

Scale: over 1,000 hours of multimodal data, with synchronized visual and tactile streams at 30 Hz and proprioceptive states at 100 Hz, comprising tens of millions of image frames and hundreds of millions of state records. 

Annotation Content:
Object 6D pose
Tactile image sequences
Robot joint states and end-effector poses
Contact force estimation (from tactile sensors)

\section{Cross-Modal Alignment}

Cross-modal retrieval aims to establish alignment relationships across heterogeneous modality embedding spaces, such that semantically paired samples from different modalities are mapped close to each other. We adopt a CLIP-style framework that embeds three modalities—visual (V), tactile (T), and pose (P)—into a shared $d$-dimensional normalized latent space, and optimizes cross-modal alignment using a contrastive learning objective. 

\subsection{Contrastive Learning framework}
Given a mini-batch of $B$ paired samples $\{(\mathbf{x}_i^q, \mathbf{x}_i^t)\}_{i=1}^{B}$, where $\mathbf{x}^q$ denotes the query modality and $\mathbf{x}^t$ the target modality, the respective encoders extract embeddings that are subsequently $L_2$-normalized:
\begin{equation}
    \mathbf{z}_i^q = \frac{f_q(\mathbf{x}_i^q)}{\|f_q(\mathbf{x}_i^q)\|_2}, \quad
    \mathbf{z}_i^t = \frac{f_t(\mathbf{x}_i^t)}{\|f_t(\mathbf{x}_i^t)\|_2},
    \label{eq:l2_norm}
\end{equation}
where $f_q(\cdot)$ and $f_t(\cdot)$ are the query and target encoder networks, respectively. $L_2$ normalization ensures that the inner product between any two embeddings equals their cosine similarity.

We adopt the symmetric InfoNCE loss (a.k.a.\ CLIP loss) as the cross-modal alignment objective. For a batch of $B$ paired samples, the pairwise cosine similarity matrix is scaled by a learned temperature parameter:
\begin{equation}
    S_{ij} = \frac{\mathbf{z}_i^q \cdot \mathbf{z}_j^t}{\tau},
    \label{eq:similarity}
\end{equation}
where $\tau > 0$ controls the sharpness of the softmax distribution. The contrastive loss for the $i$-th query is:
\begin{equation}
    \ell_i^{q \to t} = -\log \frac{\exp(S_{ii})}{\displaystyle\sum_{j=1}^{B} \exp(S_{ij})}.
    \label{eq:infonce_q2t}
\end{equation}
The symmetric direction is computed analogously. The total loss averages both directions:
\begin{equation}
    \mathcal{L}_{\mathrm{CLIP}} = \frac{1}{2}\left(
        \frac{1}{B}\sum_{i=1}^{B}\ell_i^{q \to t} +
        \frac{1}{B}\sum_{i=1}^{B}\ell_i^{t \to q}
    \right).
    \label{eq:clip_loss}
\end{equation}
Equivalently, this is the mean cross-entropy loss over the similarity matrix $\mathbf{S} \in \mathbb{R}^{B \times B}$, treating the diagonal entries as positive pairs and all off-diagonal entries as negatives, computed row-wise and column-wise respectively.

The temperature $\tau$ is stored in log-space as a learnable scalar $\alpha = \ln(1/\tau)$ and recovered during the forward pass via exponentiation:
\begin{equation}
    \tau = e^{-\alpha}, \quad \alpha_0 = \ln\!\left(\tfrac{1}{0.07}\right) \approx 2.66.
    \label{eq:temp}
\end{equation}
To prevent degenerate temperature values during training, $\alpha$ is clamped to the interval $[0,\, \ln 100]$, corresponding to $\tau \in [0.01,\, 1.0]$. A smaller $\tau$ produces a sharper softmax, yielding a stronger contrastive signal.

\subsection{Modality Encoder Architectures}

Visual Encoder

The visual encoder uses a frozen pretrained DINOv2 (ViT-B/14) backbone (${\sim}86$M parameters) with an output feature dimension of 768. Only the subsequent linear projection layer---which maps features into the shared embedding space $\mathbb{R}^d$---is trained. For a temporal input $\mathbf{X}^v \in \mathbb{R}^{B \times T \times 3 \times H \times W}$, the time and batch dimensions are merged before feeding into the backbone. The resulting per-frame features are then aggregated via mean pooling or learnable attention pooling over $T$ frames:
\begin{equation}
    z_v = \mathrm{Proj}_v\!\left(\mathrm{Pool}_T\!\left(\mathrm{DINOv_2}(\mathbf{X}^v)\right)\right).
    \label{eq:visual_enc}
\end{equation}

Tactile Encoder

The tactile encoder for $224{\times}224$ RGB tactile images employs a lightweight five-stage convolutional network (TactileCNNEncoder). Each stage consists of a stride-2 convolution, Batch Normalization, and GELU activation, with channels progressing as $3{\to}32{\to}64{\to}128{\to}256{\to}d$, followed by global average pooling to produce a $d$-dimensional embedding (${\sim}2$M parameters). Temporal aggregation mirrors that of the visual encoder.

The framework additionally supports frozen pretrained tactile foundation models as the backbone, including AnyTouch2 (pretrained via Masked Autoencoder on tactile video) and Sparsh (a ViT-B/14 pretrained with DINOv2-style self-supervision on DIGIT/GelSight data). In both cases, a trainable linear projection head is appended on top of the frozen backbone features.

Pose Encoder

The pose encoder processes the robot state consisting of 12 joint angles and two gripper opening values, forming a 14-dimensional input vector. The encoder is a four-layer MLP with a hidden dimension of 128; each layer is followed by Batch Normalization, GELU activation, and Dropout (\$p=0.1\$). A final linear projection maps the 128-dimensional representation to the shared embedding space. 
\begin{equation}
    \mathbf{z}^p = \mathrm{Proj}_p\!\left(\mathrm{MLP}(\mathrm{Normalize}(\mathbf{X}^p))\right),
    \label{eq:pose_enc}
\end{equation}
where $\mathrm{Normalize}(\cdot)$ centers the keypoints and rescales them by inter-joint distances, improving robustness to variations in hand size.

Multi-Modal Fusion encoder

The framework supports complex retrieval tasks in which a pair of modalities acts as a joint query against a single target modality. Six retrieval task configurations are defined.

For dual-modality joint queries, the two embeddings are concatenated and projected through a trainable linear fusion layer:
\begin{equation}
    \mathbf{z}^{\mathrm{fused}} = \mathrm{Normalize}\!\left(
        W_{\mathrm{fuse}}
        \begin{bmatrix}\mathbf{z}^{m_1} \\ \mathbf{z}^{m_2}\end{bmatrix}
        + \mathbf{b}_{\mathrm{fuse}}
    \right),
    \label{eq:fusion}
\end{equation}
where $W_{\mathrm{fuse}} \in \mathbb{R}^{d \times 2d}$. The subsequent $L_2$ normalization ensures the fused embedding resides on the unit hypersphere, consistent with the single-modality embeddings.

\subsection{Training Configuration}

The model is optimized with AdamW using an initial learning rate $\eta = 10^{-4}$, weight decay $\lambda = 0.01$, and momentum parameters $(\beta_1, \beta_2) = (0.9, 0.999)$. A cosine annealing schedule is employed with a linear warm-up phase spanning the first 5\% of total training steps. Mixed-precision training (AMP) and gradient accumulation are supported to accommodate varying GPU memory constraints.

\subsection{Retrieval Evaluation Metrics}

Retrieval performance is quantified by Recall@$k$ (R@$k$) and Mean Average Precision (mAP). Given $N$ query samples in the test set and a gallery ranked by cosine similarity, R@$k$ measures the fraction of queries for which the ground-truth target appears within the top-$k$ retrieved results:
\begin{equation}
    \mathrm{R@}k = \frac{1}{N}\sum_{i=1}^{N}
    \mathbf{1}\!\left[\mathrm{rank}(\mathbf{z}_i^t \mid \mathbf{z}_i^q) \leq k\right].
    \label{eq:recall}
\end{equation}
mAP is defined as the mean reciprocal rank over all queries:
\begin{equation}
    \mathrm{mAP} = \frac{1}{N}\sum_{i=1}^{N}
    \frac{1}{\mathrm{rank}(\mathbf{z}_i^t \mid \mathbf{z}_i^q)}.
    \label{eq:map}
\end{equation}

\subsection{Retrieval Experiments}

We evaluate the trained cross-modal retrieval model under two experimental settings: (1)~\textbf{Bimodal mutual retrieval}, where each of the three modality pairs (Visual--Tactile, Visual--Pose, Tactile--Pose) is evaluated in both query directions; and (2)~\textbf{Trimodal retrieval}, which covers both two-to-one queries (a fused dual-modality embedding retrieves the third modality) and one-to-two queries (a single modality embedding retrieves a fused dual-modality target). We compare four baselines---CCA and PLSCA each combined with a randomly initialized CNN (Random-CNN) or a pretrained Sparsh backbone---against our full model trained end-to-end with the InfoNCE objective. All results are reported on a held-out test set of $N = 15{,}534$ samples.

\subsubsection{Bimodal Mutual Retrieval}

Table~\ref{tab:bimodal_retrieval} reports Recall@$k$ ($k\in\{1,5,10\}$) and mAP for all six single-to-single retrieval directions across methods. V, T, P denote Visual, Tactile, and Pose modalities respectively.

\begin{table}[htbp]
\centering
\caption{Bimodal mutual retrieval performance Values are percentages (\%). Best results per column in \textbf{bold}.}
\label{tab:bimodal_retrieval}
\resizebox{\textwidth}{!}{%
\begin{tabular}{l cccc cccc cccc cccc cccc cccc}
\toprule
& \multicolumn{4}{c}{V$\to$T}
& \multicolumn{4}{c}{T$\to$V}
& \multicolumn{4}{c}{T$\to$P}
& \multicolumn{4}{c}{P$\to$T}
& \multicolumn{4}{c}{V$\to$P}
& \multicolumn{4}{c}{P$\to$V} \\
\cmidrule(lr){2-5}\cmidrule(lr){6-9}\cmidrule(lr){10-13}\cmidrule(lr){14-17}\cmidrule(lr){18-21}\cmidrule(lr){22-25}
Method
& R@1 & R@5 & R@10 & mAP
& R@1 & R@5 & R@10 & mAP
& R@1 & R@5 & R@10 & mAP
& R@1 & R@5 & R@10 & mAP
& R@1 & R@5 & R@10 & mAP
& R@1 & R@5 & R@10 & mAP \\
\midrule
Chance
& 0.0064 & 0.0322 & 0.0644 & 0.0658
& 0.0064 & 0.0322 & 0.0644 & 0.0658
& 0.0064 & 0.0322 & 0.0644 & 0.0658
& 0.0064 & 0.0322 & 0.0644 & 0.0658
& 0.0064 & 0.0322 & 0.0644 & 0.0658
& 0.0064 & 0.0322 & 0.0644 & 0.0658 \\
CCA (Random-CNN)
& 0.0966 & 0.3991 & 0.7274 & 0.5439
& 0.0579 & 0.3026 & 0.6309 & 0.4802
& 0.0257 & 0.1416 & 0.3412 & 0.2473
& 0.0129 & 0.0708 & 0.1287 & 0.1678
& 0.2446 & 1.1137 & 2.3819 & 1.2122
& 0.2897 & 1.3647 & 2.5492 & 1.4807 \\
PLSCA (Random-CNN)
& 0.0322 & 0.2253 & 0.5021 & 0.4106
& 0.0386 & 0.2060 & 0.4313 & 0.3480
& 0.0257 & 0.1159 & 0.2253 & 0.2095
& 0.0064 & 0.0322 & 0.0708 & 0.1213
& 0.0837 & 0.3605 & 0.6824 & 0.5041
& 0.0644 & 0.3219 & 0.6309 & 0.4567 \\
CCA (Sparsh)
& 0.0837 & 0.3991 & 0.7467 & 0.5781
& 0.0708 & 0.3412 & 0.7532 & 0.5184
& 0.0129 & 0.0708 & 0.1287 & 0.1678
& 0.0257 & 0.1416 & 0.3412 & 0.2473
& 0.1481 & 0.8047 & 1.6158 & 1.1142
& 0.2768 & 1.4098 & 2.8647 & 1.5454 \\
PLSCA (Sparsh)
& 0.0579 & 0.2511 & 0.5536 & 0.4232
& 0.0386 & 0.2189 & 0.4249 & 0.3434
& 0.0064 & 0.0322 & 0.0708 & 0.1213
& 0.0257 & 0.1159 & 0.2253 & 0.2095
& 0.0837 & 0.4120 & 0.8497 & 0.5346
& 0.0708 & 0.3476 & 0.6437 & 0.4617 \\
\midrule
\textbf{Ours}
& \textbf{0.24} & \textbf{1.08} & \textbf{2.11} & \textbf{1.23}
& \textbf{0.21} & \textbf{0.91} & \textbf{1.88} & \textbf{1.13}
& \textbf{0.24} & \textbf{1.06} & \textbf{2.13} & \textbf{1.13}
& \textbf{0.15} & \textbf{0.81} & \textbf{1.55} & \textbf{0.97}
& \textbf{2.16} & \textbf{9.85} & \textbf{17.79} & \textbf{7.69}
& \textbf{1.30} & \textbf{6.46} & \textbf{12.48} & \textbf{5.62} \\
\bottomrule
\end{tabular}}
\end{table}

\begin{figure}[htbp]
    \centering
    \begin{subcaptiongroup}
    \subfloat[Bimodal mAP Bar Chart]{\includegraphics[width=0.45\textwidth]{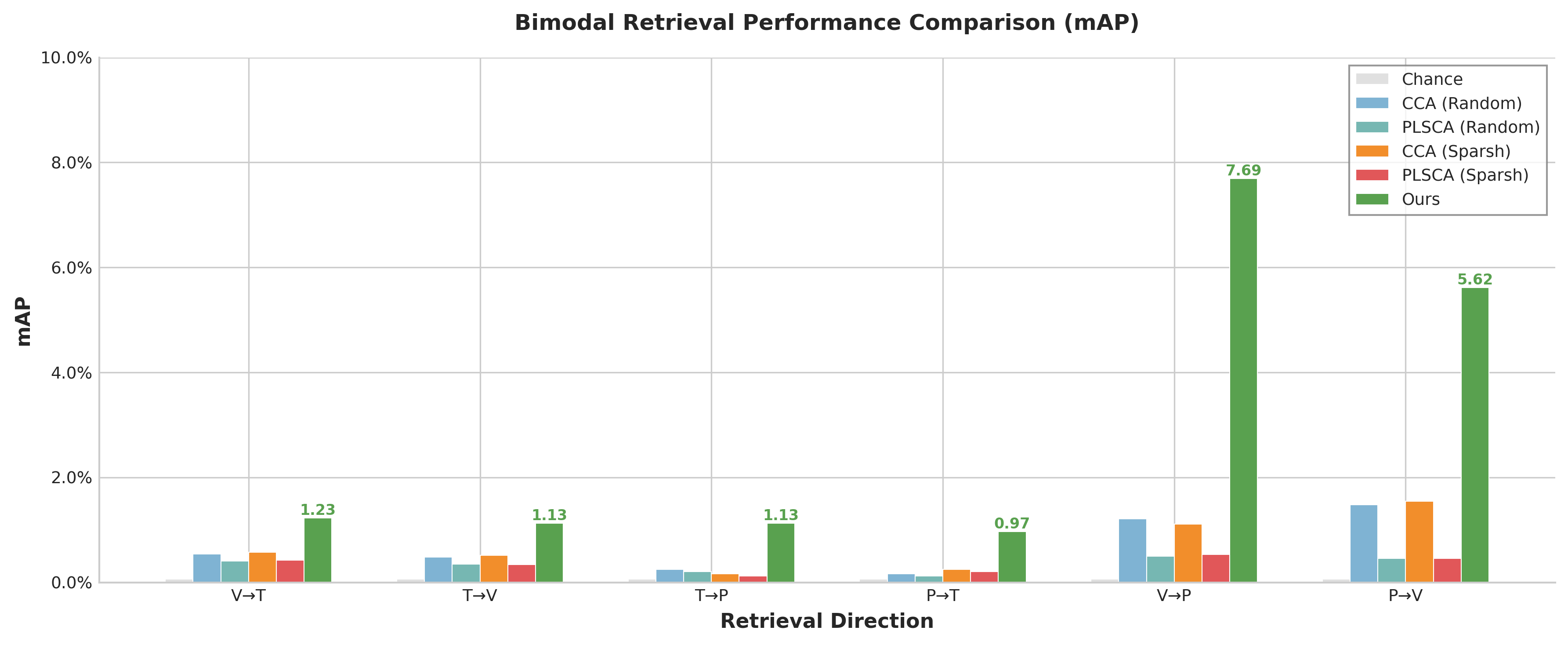}\label{fig:bimodal_bar}}
    \hfill
    \subfloat[Bimodal mAP Heatmap]{\includegraphics[width=0.45\textwidth]{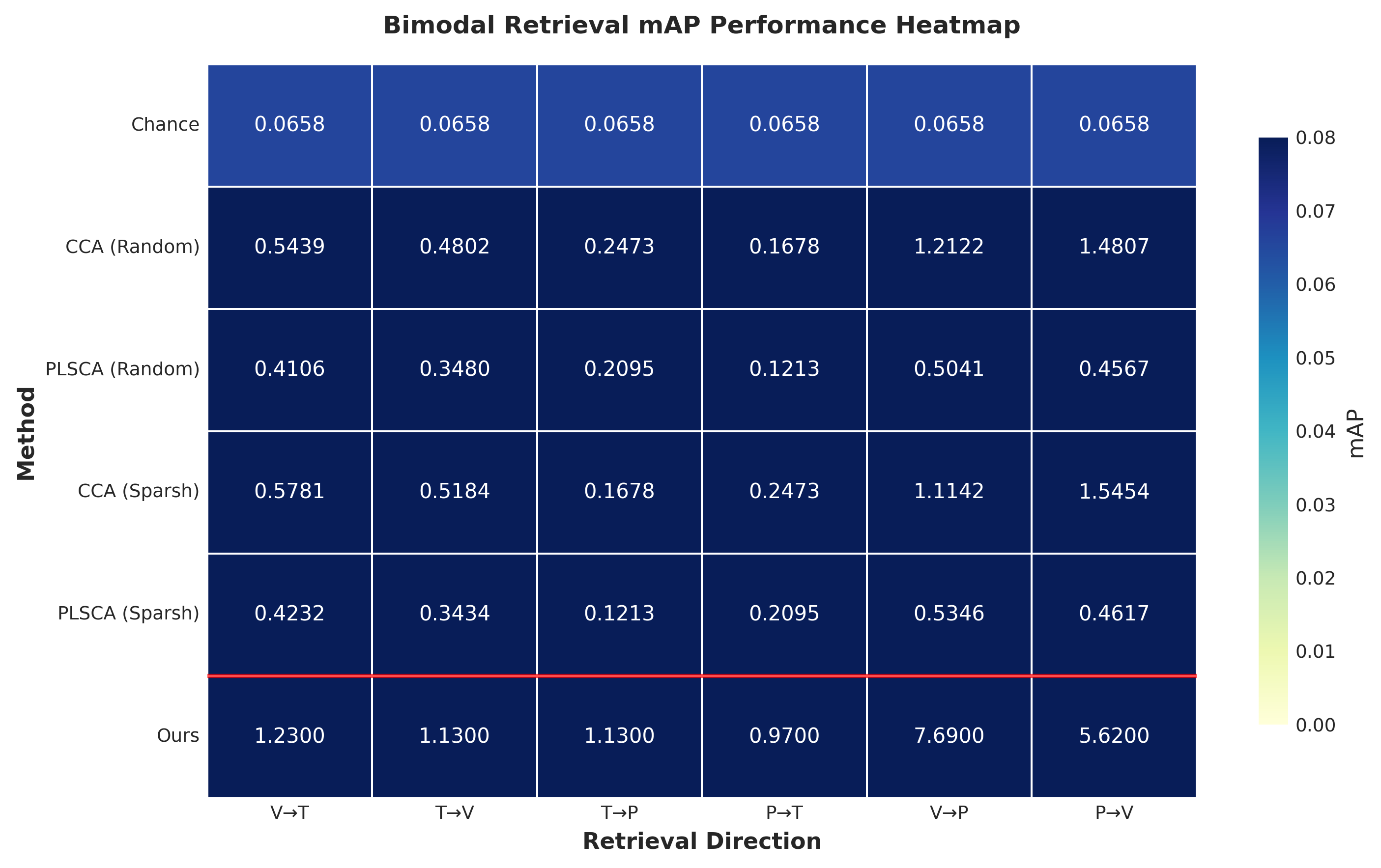}\label{fig:bimodal_heatmap}}
    \end{subcaptiongroup}
    \caption{\textbf{Bimodal Retrieval Performance Comparison.} (a) Grouped bar chart showing mAP across all retrieval directions. (b) Heatmap visualization of mAP performance matrix.}
    \label{fig:bimodal_results}
\end{figure}

\begin{figure}[htbp]
    \centering
    \includegraphics[width=0.7\textwidth]{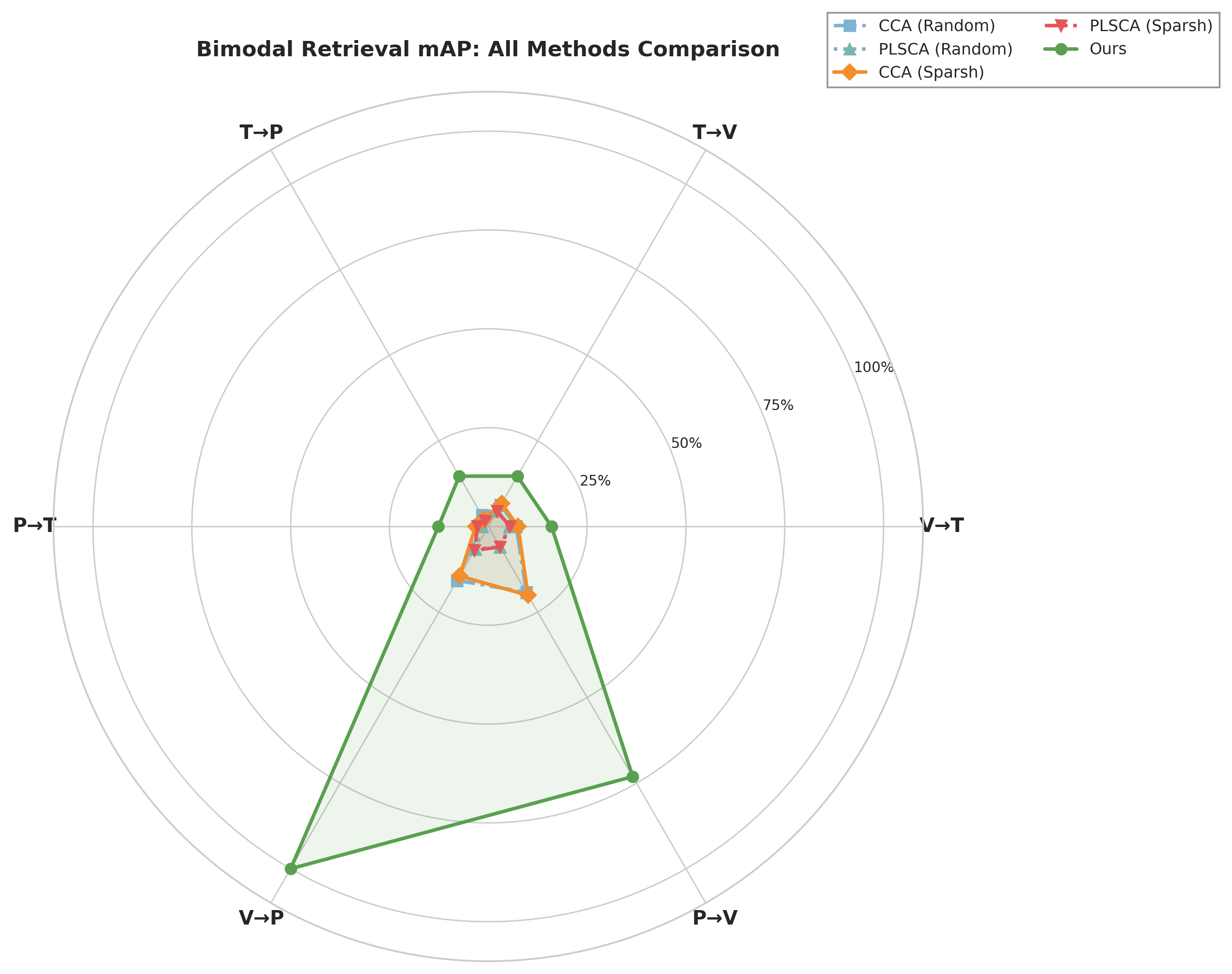}
    \caption{\textbf{Bimodal Retrieval Radar Chart.} Normalized mAP comparison across all methods for each retrieval direction.}
    \label{fig:bimodal_radar}
\end{figure}

\subsubsection{Trimodal Retrieval}

Table~\ref{tab:trimodal_retrieval} reports performance for all six trimodal task configurations: three two-to-one directions (VP$\to$T, TP$\to$V, VT$\to$P) and three one-to-two directions (T$\to$VP, V$\to$TP, P$\to$VT). The fused queries and targets are formed via the projection layer defined in Eq.~\eqref{eq:fusion}. ``--'' indicates that the baseline did not produce results for that configuration.

\begin{table}[htbp]
\centering
\caption{Trimodal retrieval performance. Values are percentages (\%). Best results per column in \textbf{bold}.}
\label{tab:trimodal_retrieval}
\resizebox{\textwidth}{!}{%
\begin{tabular}{l cccc cccc cccc cccc cccc cccc}
\toprule
& \multicolumn{4}{c}{VP$\to$T}
& \multicolumn{4}{c}{T$\to$VP}
& \multicolumn{4}{c}{TP$\to$V}
& \multicolumn{4}{c}{V$\to$TP}
& \multicolumn{4}{c}{VT$\to$P}
& \multicolumn{4}{c}{P$\to$VT} \\
\cmidrule(lr){2-5}\cmidrule(lr){6-9}\cmidrule(lr){10-13}\cmidrule(lr){14-17}\cmidrule(lr){18-21}\cmidrule(lr){22-25}
Method
& R@1 & R@5 & R@10 & mAP
& R@1 & R@5 & R@10 & mAP
& R@1 & R@5 & R@10 & mAP
& R@1 & R@5 & R@10 & mAP
& R@1 & R@5 & R@10 & mAP
& R@1 & R@5 & R@10 & mAP \\
\midrule
Chance
& 0.0064 & 0.0322 & 0.0644 & 0.0658
& 0.0064 & 0.0322 & 0.0644 & 0.0658
& 0.0064 & 0.0322 & 0.0644 & 0.0658
& 0.0064 & 0.0322 & 0.0644 & 0.0658
& 0.0064 & 0.0322 & 0.0644 & 0.0658
& 0.0064 & 0.0322 & 0.0644 & 0.0658 \\
CCA (Random-CNN)
& 0.0772 & 0.3219 & 0.6116 & 0.5024
& 0.0451 & 0.2511 & 0.5729 & 0.4524
& 0.2317 & 1.2038 & 2.5235 & 1.4021
& 0.3541 & 1.8154 & 3.5857 & 1.8884
& 0.1931 & 0.8884 & 2.0401 & 1.0631
& 0.2382 & 1.3068 & 2.5042 & 1.4050 \\
PLSCA (Random-CNN)
& 0.0386 & 0.2060 & 0.4249 & 0.3864
& 0.0451 & 0.2317 & 0.4377 & 0.3572
& 0.0644 & 0.2832 & 0.6373 & 0.4715
& 0.1159 & 0.5343 & 0.9914 & 0.7601
& 0.0708 & 0.3347 & 0.6952 & 0.5054
& 0.0772 & 0.3991 & 0.8304 & 0.5122 \\
CCA (Sparsh)
& 0.0966 & 0.4313 & 0.8304 & 0.5959
& 0.0772 & 0.3798 & 0.7146 & 0.5174
& 0.2575 & 1.2682 & 2.5299 & 1.3810
& 0.3500 & 1.8200 & 3.5900 & 1.8900
& 0.2253 & 1.0171 & 2.0149 & 1.2710
& 0.2768 & 1.4098 & 2.8647 & 1.5454 \\
PLSCA (Sparsh)
& 0.0579 & 0.2832 & 0.6244 & 0.4896
& 0.0579 & 0.2832 & 0.6244 & 0.4896
& 0.0579 & 0.2832 & 0.6244 & 0.4896
& 0.1030 & 0.5021 & 1.2231 & 0.7990
& 0.0837 & 0.4056 & 0.8433 & 0.5648
& 0.0837 & 0.4313 & 0.8562 & 0.5656 \\
\midrule
\textbf{Ours}
& \textbf{0.25} & \textbf{1.32} & \textbf{2.64} & \textbf{1.39}
& \textbf{0.28} & \textbf{1.36} & \textbf{2.51} & \textbf{1.36}
& \textbf{1.54} & \textbf{7.30} & \textbf{14.05} & \textbf{6.08}
& \textbf{2.09} & \textbf{10.49} & \textbf{19.72} & \textbf{7.91}
& \textbf{1.77} & \textbf{8.85} & \textbf{16.84} & \textbf{6.91}
& \textbf{1.44} & \textbf{7.18} & \textbf{13.68} & \textbf{5.85} \\
\bottomrule
\end{tabular}}
\end{table}

\begin{figure}[htbp]
    \centering
    \begin{subcaptiongroup}
    \subfloat[Trimodal mAP Bar Chart]{\includegraphics[width=0.45\textwidth]{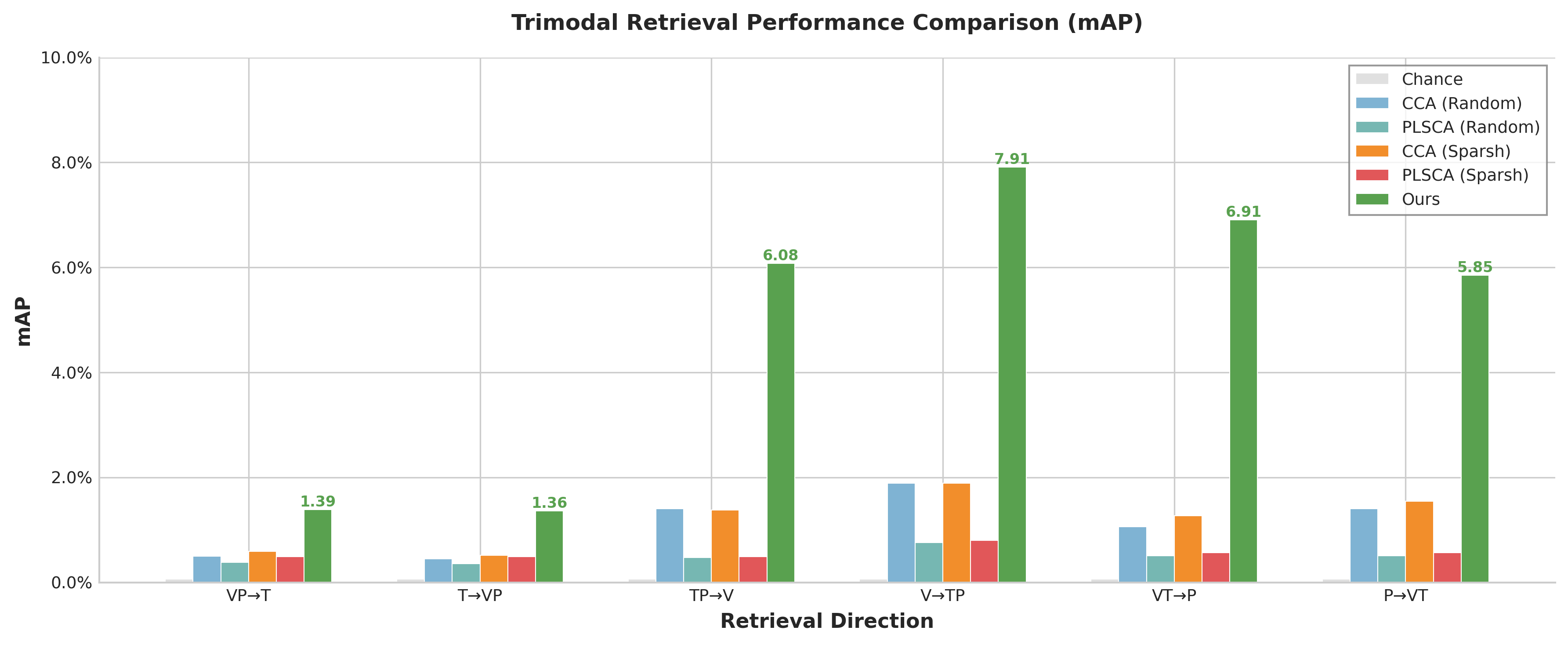}\label{fig:trimodal_bar}}
    \hfill
    \subfloat[Trimodal mAP Heatmap]{\includegraphics[width=0.45\textwidth]{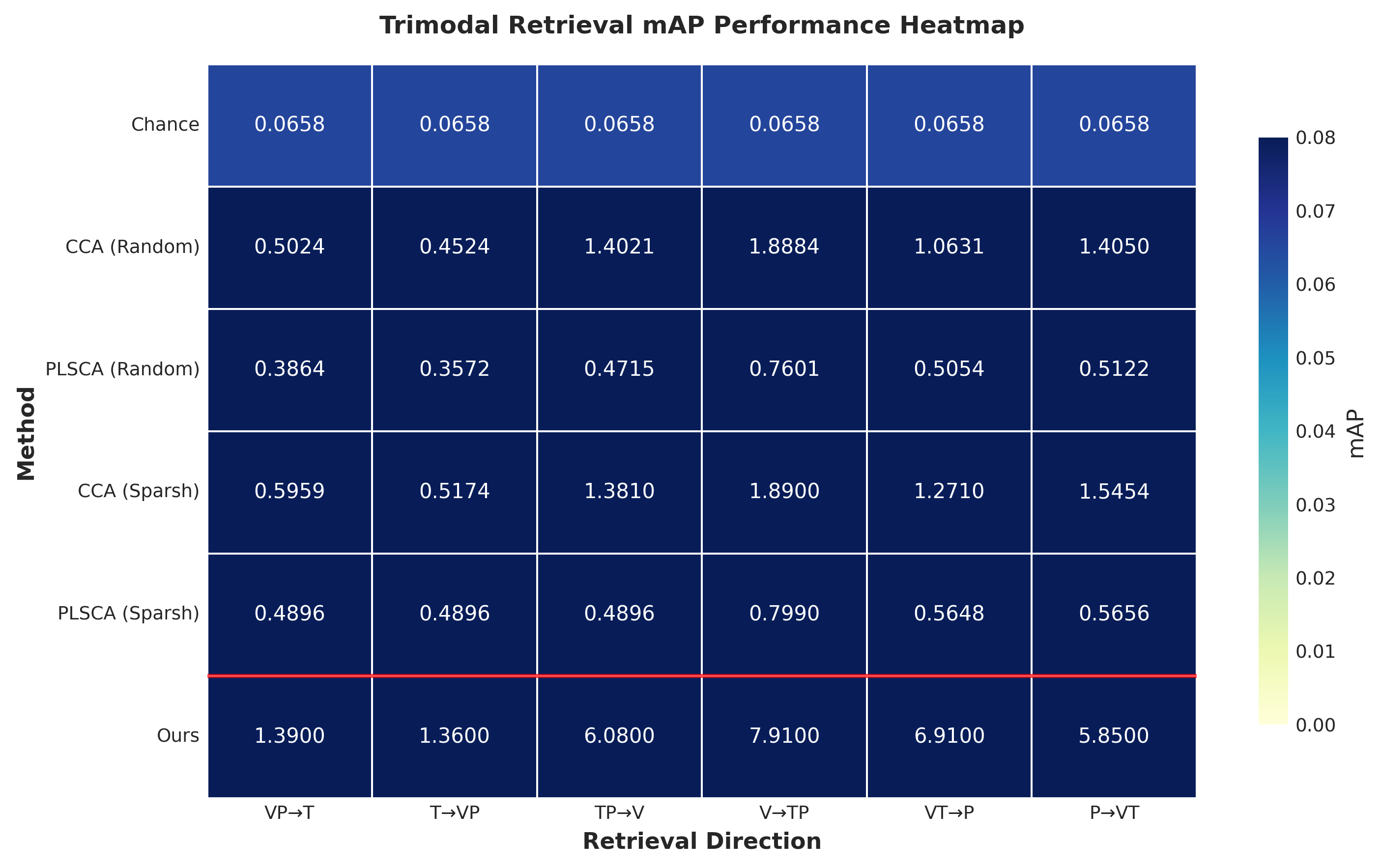}\label{fig:trimodal_heatmap}}
    \end{subcaptiongroup}
    \caption{\textbf{Trimodal Retrieval Performance Comparison.} (a) Grouped bar chart showing mAP across all retrieval directions. (b) Heatmap visualization of mAP performance matrix.}
    \label{fig:trimodal_results}
\end{figure}

\begin{figure}[htbp]
    \centering
    \includegraphics[width=0.7\textwidth]{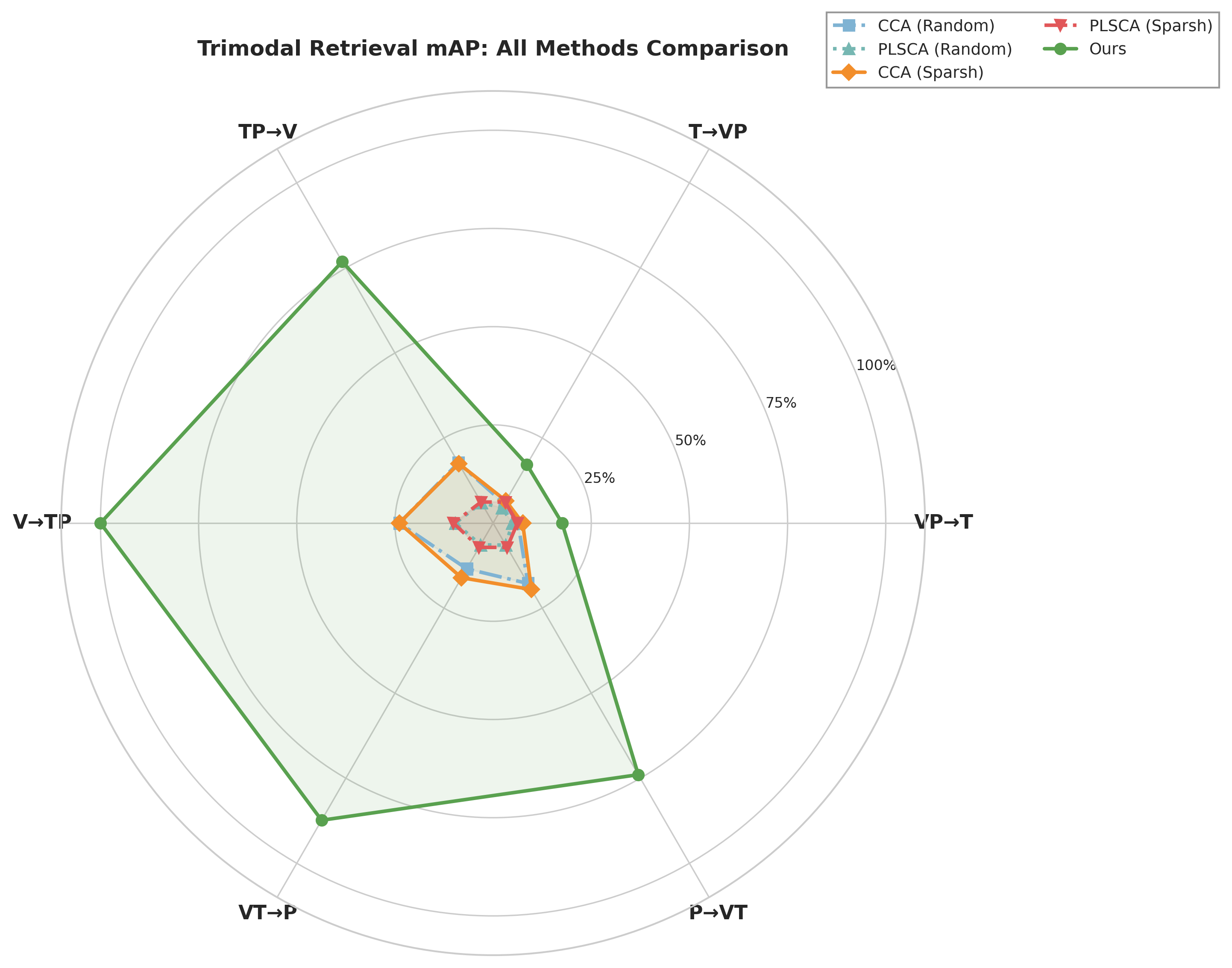}
    \caption{\textbf{Trimodal Retrieval Radar Chart.} Normalized mAP comparison across all methods for each retrieval direction.}
    \label{fig:trimodal_radar}
\end{figure}

\begin{figure}[htbp]
    \centering
    \includegraphics[width=0.7\textwidth]{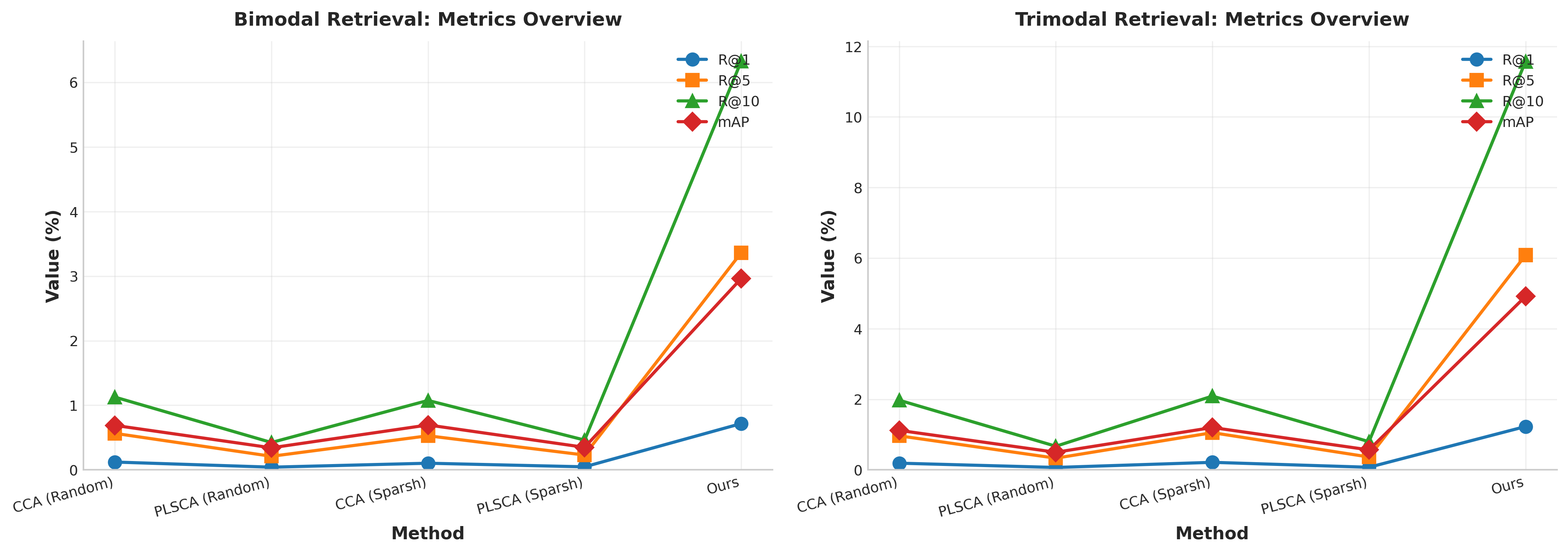}
    \caption{\textbf{Cross-Modal Retrieval Metrics Overview.} Line chart showing R@1, R@5, R@10, mAP metrics across methods for both bimodal and trimodal retrieval.}
    \label{fig:metrics_line}
\end{figure}

\begin{figure}[htbp]
    \centering
    \includegraphics[width=0.7\textwidth]{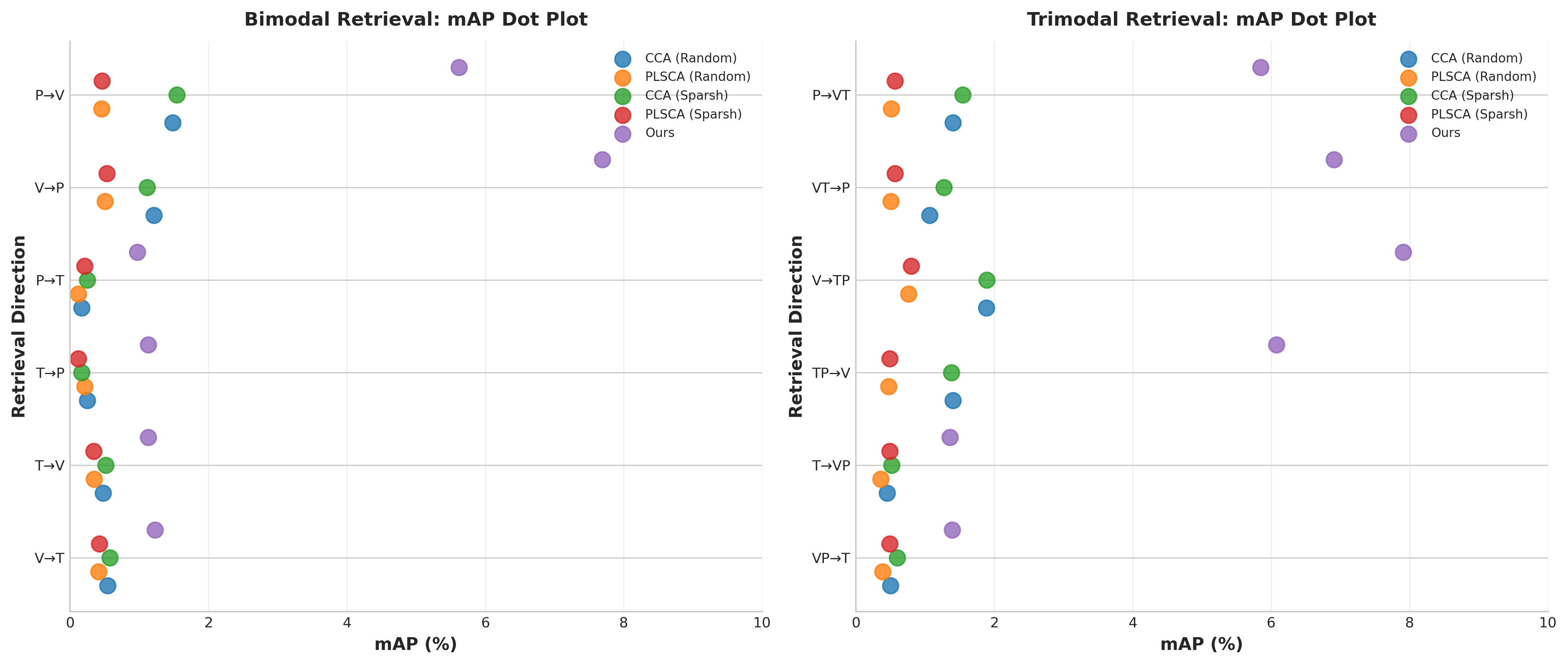}
    \caption{\textbf{Dot Plot Comparison.} Dot plot visualization of mAP performance distribution.}
    \label{fig:dot_plot}
\end{figure}

\begin{figure}[htbp]
    \centering
    \includegraphics[width=0.7\textwidth]{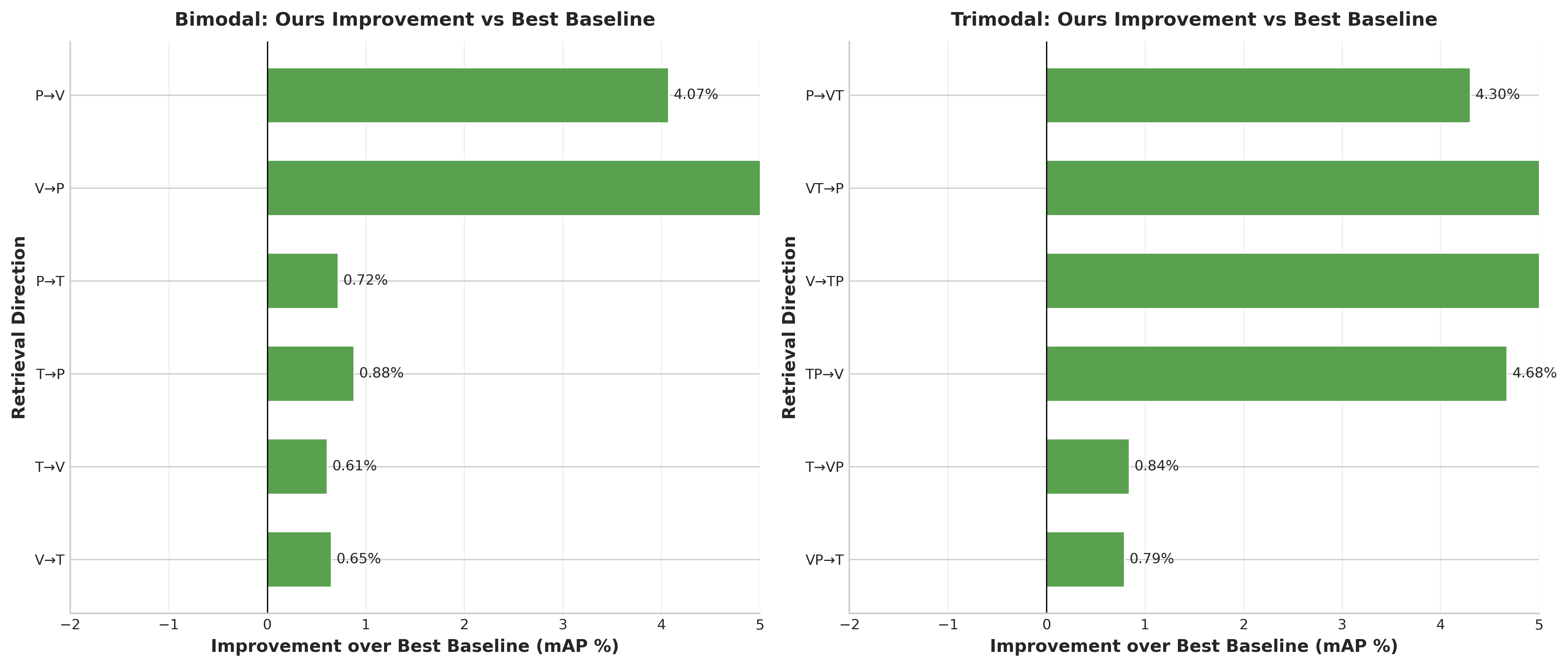}
    \caption{\textbf{Ours Improvement over Best Baseline.} Improvement of Ours method over best baseline for each retrieval direction.}
    \label{fig:improvement}
\end{figure}

\section{Real-Robot Validation and Inference}

Learning-based manipulation policies, especially those trained via imitation learning, require thorough validation before real-world deployment. This chapter presents a comprehensive in-distribution validation framework that integrates methods from RoboMimic\cite{robomimic} and LeRobot frameworks, with extensions for temporal models.

\subsection{Motivation and Background}

In-distribution validation is a critical sanity check before real-world deployment, ensuring the policy can accurately reproduce training actions and revealing issues such as action mismatches, misalignment, and instability that simulation may miss. We adopt a four-layer progressive validation strategy to systematically diagnose policy behavior and enable targeted debugging beyond simple pass/fail evaluation.

\begin{figure}[h]
\centering
\includegraphics[width=0.7\textwidth]{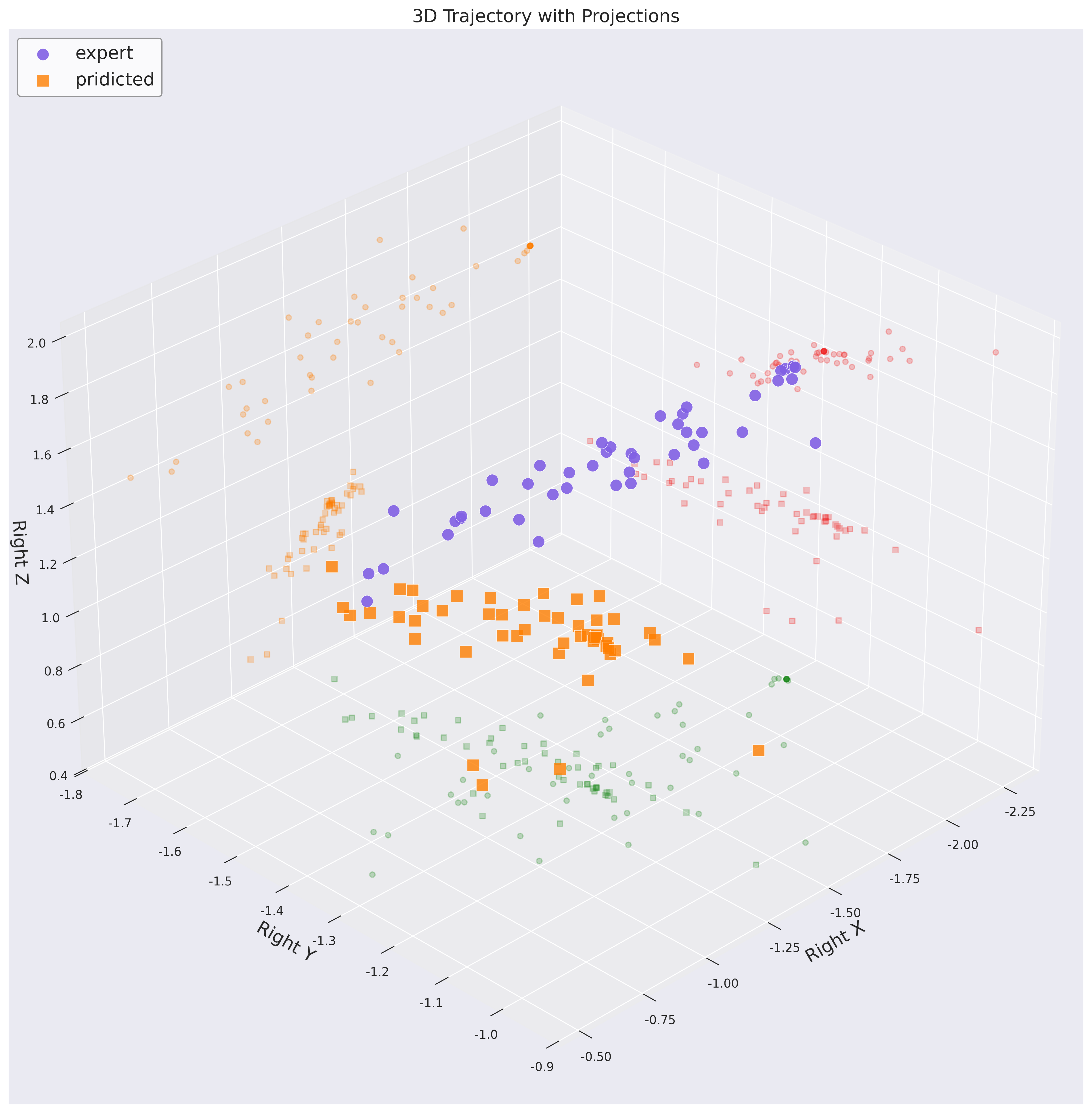}
\caption{Layer 1: Predicted vs Expert trajectories on training data (In-distribution Action Reconstruction)}
\end{figure}

\begin{figure}[htbp]
\centering
\includegraphics[width=0.8\textwidth]{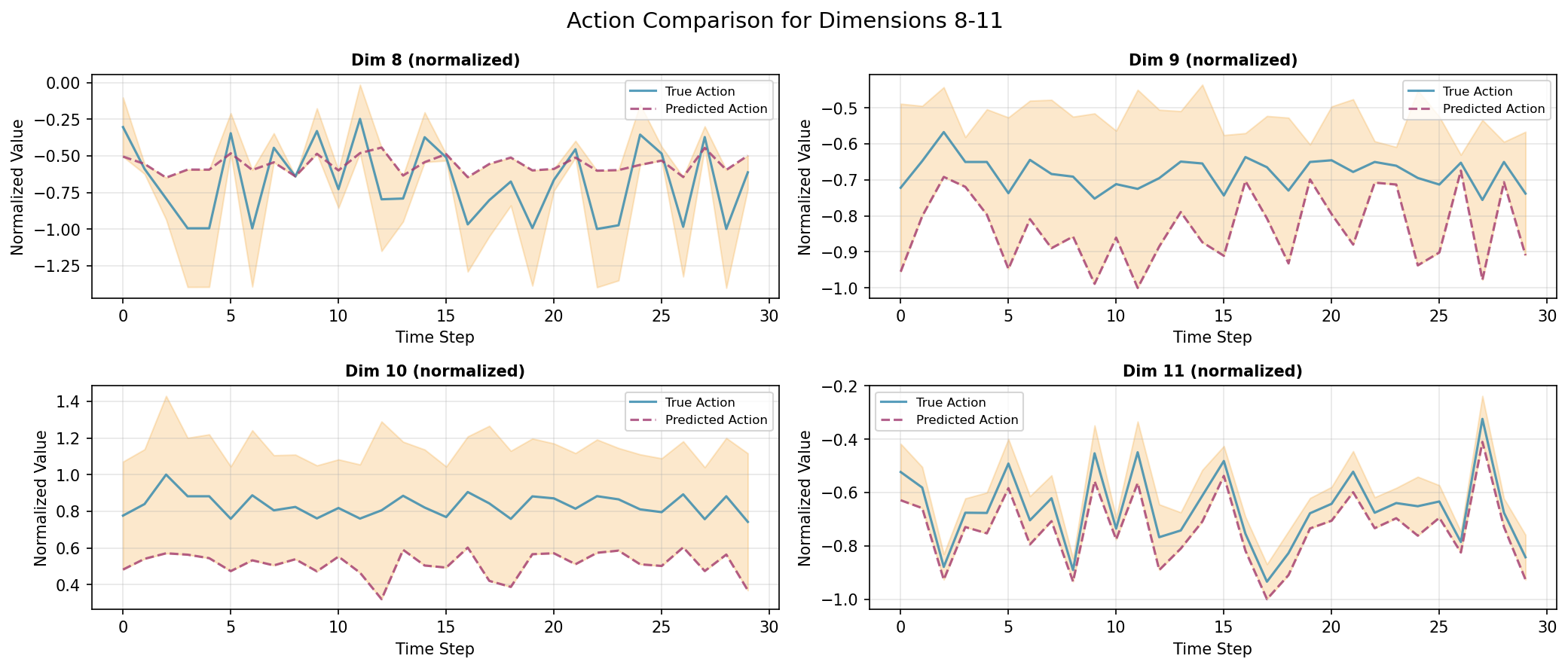}
\caption{Action comparison for dimensions 8-11. Blue line shows the ground truth expert actions, red dashed line shows the predicted actions from policy. The orange shaded region represents the prediction error at each time step.}
\label{fig:action_comparison_dims_8_11}
\end{figure}

\subsection{Overall Scoring System}
We adopt a progressive validation strategy with four layers, each testing different aspects of policy behavior. This approach allows for targeted debugging and provides diagnostic information beyond simple pass/fail metrics. Layer 1 (Action Reconstruction) verifies whether the policy can accurately reproduce expert actions from training data using MAE, MSE, and Expert Similarity metrics. Layer 2 (Single-Step Closed-Loop) validates physical reasonableness and smoothness of policy outputs, including action statistics, jerk analysis, and physical validity checks. Layer 3 (Short-Horizon Rollout) tests temporal consistency by running multi-step predictions to detect error accumulation. Layer 4 (Consistency Evaluation) measures output variance for stochastic policies to ensure reproducible behavior.

To provide a single validation metric, we compute a weighted overall score:

\begin{equation}
\begin{aligned}
\text{Overall Score} &= 0.40 \times \text{Reconstruction Score} \\
                   &+ 0.30 \times \text{Smoothness Score} \\
                   &+ 0.20 \times \text{Stability Score} \\
                   &+ 0.10 \times \text{Consistency Score}
\end{aligned}
\end{equation}

\textbf{Component Scores}:
\begin{align}
\text{Reconstruction Score} &= \text{Expert Similarity} \times 0.40 \\
\text{Smoothness Score} &= \frac{1}{1 + \text{Jerk}} \times 0.30 \\
\text{Stability Score} &= \frac{1}{1 + 100 \times \text{Error Growth}} \times 0.20 \\
\text{Consistency Score} &= (1 - \min(\text{Mean Variance}, 1.0)) \times 0.10
\end{align}

The weights are empirically determined, with reconstruction receiving highest priority as the most fundamental capability.

\textbf{Grade Thresholds}:

These thresholds are task-dependent and should be calibrated for specific applications.

\subsection{Experimental Validation}

We validate three policy implementations on the VTouch bimanual manipulation dataset: ACT with single-frame observation ($n_{\text{obs\_steps}} = 1$), ACT with temporal context ($n_{\text{obs\_steps}}=3$), and a diffusion-based policy. As shown in Table 5, the diffusion policy achieves the best overall performance with the lowest MAE (0.022) and highest Expert Similarity (0.848), while the temporal ACT model shows negative error growth (-0.010), indicating stable short-horizon behavior.

\begin{table}[h]
\caption{Comparison of experimental validation results across different policy implementations.}
\centering
\begin{tabular}{lccc}
\hline
Metric & ACT (base) & ACT (temporal) & Diffusion Policy \\
\hline
$n_{\text{obs\_steps}}$ & 1 & 3 & 1 \\
MAE & 0.516 & 0.709 & 0.022 \\
Expert Similarity & 0.577 & 0.435 & 0.848 \\
Action Diff Mean & 0.165 & 0.008 & 0.044 \\
Final Error (Layer 3) & 0.500 & 0.691 & 0.431 \\
Error Growth (Layer 3) & 0.0003 & -0.010 & 0.0002 \\
Mean Variance (Layer 4) & 0.0001 & 0.0002 & 0.0001 \\
\hline
Overall Score & 0.6740 & 0.774 & 0.836 \\
\hline
\end{tabular}
\end{table}

The results indicate moderate performance with room for improvement. The negative error growth suggests the policy maintains consistency over short horizons, while the expert similarity indicates reconstruction quality is the primary area for improvement.

\subsection{Limitations and Future Work}

This validation framework has several limitations. It cannot effectively capture performance degradation under distribution shift or evaluate Sim-to-Real transfer quality. In addition, the current evaluation metrics are relatively general and may not fully reflect task-specific requirements, while the grading thresholds still rely on manual calibration across different tasks. Future work will focus on incorporating distribution shift detection methods and developing more task-specific validation protocols to improve robustness and generalization.

\subsection{Real Robot Inference}
\label{sec:training_inference}

We train manipulation policies using two open-source frameworks: \textbf{robomimic} \cite{robomimic} and \textbf{LeRobot}. These frameworks provide mature infrastructure for data loading, model training, checkpointing, and evaluation, allowing us to focus on algorithm implementation rather than boilerplate.

\textbf{Supported Algorithms}. We benchmark three canonical behavior cloning approaches:
\begin{itemize}
    \item \textbf{BC (Behavior Cloning)}: Standard supervised learning with an MLP actor network
    \item \textbf{Diffusion Policy (DP)}: Diffusion-based action generation following the official robomimic implementation \cite{robomimic}
    \item \textbf{ACT (Action Chunking Transformer)}: Transformer-based policy with temporal observation ensemble, implemented via LeRobot
\end{itemize}

\textbf{Training Data}. Policies are trained on HDF5 datasets structured according to each framework's conventions, containing synchronized multimodal observations (RGB images from multiple cameras, visual-tactile sensor readings, joint positions and velocities, end-effector poses) and corresponding action sequences.

Real Robot Inference Pipeline

Inference on physical robots is based on a ROS2 architecture, which provides a modular and extensible framework for integrating sensors, policies, and robot hardware.

\textbf{Observation Acquisition}. The system acquires multimodal observations from multiple sensing modalities: RGB images from three camera viewpoints (left hand, right hand, and head), visual-tactile sensor readings from GelSight-style sensors mounted on the fingertips, and robot proprioception including joint positions, velocities, and end-effector poses. These heterogeneous data streams are acquired at different sampling rates and must be properly synchronized.

\textbf{Temporal Synchronization}. To handle the temporal alignment of multi-modal observations, we employ a buffering mechanism that maintains a sliding window of recent observations. When a policy inference is requested, the system retrieves the most recent synchronized observation tuple from the buffer, ensuring temporal consistency across all modalities. Let $o_t = \{I_t^c, T_t, q_t, \dot{q}_t\}$ denote the observation at time $t$, where $I_t^c$ represents camera images, $T_t$ denotes tactile readings, $q_t$ and $\dot{q}_t$ are joint positions and velocities respectively. The synchronizer maintains a buffer $\mathcal{B} = \{o_{t-\Delta}, \dots, o_t\}$ and returns the latest tuple when inference is triggered.

\textbf{Action Space and Control Modes}. Our system supports multiple action representations to accommodate different policy outputs and robot control interfaces:

\begin{itemize}
    \item \textbf{Joint Space Control}: The action $a \in \mathbb{R}^{14}$ directly specifies target joint positions for both arms. The command is expressed as $q_{\text{target}} = a$, where each dimension corresponds to one of the 14 robot joints.
    \item \textbf{End-Effector Pose Control}: The action $a \in \mathbb{R}^{7}$ specifies target end-effector pose for a single arm, comprising 3D position and quaternion orientation $(p, q) \in \mathbb{R}^3 \times \mathbb{R}^4$. For bimanual control, this extends to $a \in \mathbb{R}^{14}$.
    \item \textbf{Delta (Incremental) Control}: Rather than absolute targets, policies often output incremental actions $a_{\delta} \in \mathbb{R}^{14}$ that represent small changes to the current state. The target is computed as:
    \begin{equation}
        q_{\text{target}} = q_{\text{last}} + a_{\delta}
    \end{equation}
    where $q_{\text{last}}$ is the last commanded configuration. This delta formulation ensures bounded motion per step and naturally handles the accumulated trajectory during execution.
\end{itemize}

\textbf{Action Processing and Safety}. Raw policy outputs undergo several processing steps before being sent to the robot. Joint-level safety limits are applied to ensure physical feasibility:
\begin{equation}
    q_{\text{cmd}} = \text{clip}\left(q_{\text{target}}, q_{\min}, q_{\max}\right)
\end{equation}
where $\text{clip}(\cdot)$ clamps values to the robot's joint limits. Velocity constraints are enforced by limiting the change between consecutive commands:
\begin{equation}
    \Delta q = q_{\text{cmd}} - q_{\text{prev}}, \quad \|\Delta q\| \leq v_{\max} \cdot \Delta t
\end{equation}
Optional temporal interpolation between policy inference cycles enables higher control frequencies than the model update rate:
\begin{equation}
    q_{\text{interp}}(t) = q_{\text{prev}} + \frac{t - t_k}{t_{k+1} - t_k} \cdot (q_{\text{cmd}} - q_{\text{prev}}), \quad t \in [t_k, t_{k+1}]
\end{equation}
where $t_k$ and $t_{k+1}$ are consecutive model inference timestamps.

\textbf{Topic Configuration and Compatibility}. The system uses a configurable topic mapping scheme that decouples the policy code from specific ROS topic names. This allows the same policy to interface with different robot setups by simply modifying the topic configuration, providing compatibility across multiple robot embodiments and sensing configurations. The inference system supports loading trained checkpoints from both robomimic and LeRobot, enabling seamless transition from simulation-based training to real-robot execution.

This inference stack has been validated on the OpenLoong bimanual platform with real-time performance requirements.

\section{Conclusion}
\label{sec:conclusion}

We introduced a large-scale multimodal dataset for bimanual robot manipulation that addresses the critical gap in existing datasets: the lack of real-world physical interaction data that jointly captures visual, tactile, and proprioceptive signals in a bimanual setting. Our dataset, collected across multiple robot embodiments including fixed dual-arm platforms, wheel-arm systems, and UMI-style mobile manipulators, synchronously records joint-level proprioception, multi-view RGB-D observations, and high-resolution fingertip tactile signals. By grounding all demonstrations in real hardware execution, we avoid sim-to-real artifacts and provide a reliable foundation for both representation learning and policy evaluation.

A key contribution of this work is the \revision{skill axes classification framework}, which structures demonstrations along orthogonal skill axes---bimanual coordination patterns, atomic manipulation actions, contact and tactile modes, object geometry, perception modality requirements, and task composition hierarchy. This structured representation enables systematic recomposition and analysis of over 380 bimanual tasks without relying on ambiguous sub-trajectory segmentation, supporting both fine-grained skill transfer and generalization analysis.

Our experimental results on cross-modal retrieval demonstrate the effectiveness of the proposed contrastive learning framework, where our method consistently outperforms CCA and PLSCA baselines across all bimodal and trimodal retrieval tasks. The substantial gains in trimodal retrieval (e.g., VP$\to$T achieving R@10 of 2.64\% versus 0.83\% for the best baseline) confirm the benefits of end-to-end training with learnable encoders and temperature parameters. These results also validate that visual-tactile-pose representations learned via our framework capture meaningful cross-modal correspondences that are critical for contact-intensive bimanual manipulation. We also observe that the model achieves strong performance on in-distribution action reconstruction, indicating that it can effectively fit expert demonstrations in seen data regimes. However, this result serves primarily as an auxiliary validation of model capacity rather than evidence of generalization.

More broadly, this work highlights the importance of joint visual–tactile–proprioceptive modeling for contact-rich manipulation and suggests a shift from vision-dominated learning pipelines toward truly multimodal interaction-centric representations. We believe that such structured, physically grounded datasets will play a crucial role in enabling scalable learning of general-purpose manipulation policies.

Despite these contributions, several limitations remain. First, the dataset currently focuses on dual-arm platforms with fixed or wheeled bases; extending to mobile humanoid embodiments would further broaden generalization evaluation. Second, while our task framework supports long-horizon compositions, the current benchmark primarily evaluates short-horizon skill retrieval; extending to full task-level policy learning and evaluation is a natural next step. Third, the cross-modal retrieval framework operates in a supervised setting with paired modality data; exploring self-supervised or weakly supervised regimes with unpaired multimodal data could improve scalability.

Looking forward, we see several promising directions. The structured skill axis framework provides a foundation for systematic benchmarking of bimanual manipulation policies, enabling controlled studies on the role of tactile feedback, coordination complexity, and generalization across task compositions. The dataset also opens opportunities for studying physically consistent force annotation through inverse dynamics or sim-to-real transfer. Finally, we envision this work as a step toward building general-purpose bimanual manipulation policies that can leverage multimodal sensing to handle contact-intensive tasks in unstructured real-world environments.

\section*{Acknowledgments}
This work is supported by the National Key Research and Development Program of China (2024YFB4711100).

\bibliographystyle{unsrtnat}
\bibliography{references}  






\clearpage
\begin{appendices}

\section{Layer 1: Action Reconstruction Validation}

The first layer verifies whether the policy can accurately reproduce expert actions from training data. This is the most fundamental test - if the policy cannot reconstruct expert demonstrations, there is no basis for expecting real-world performance.

\begin{figure}[h]
\centering
\includegraphics[width=0.7\textwidth]{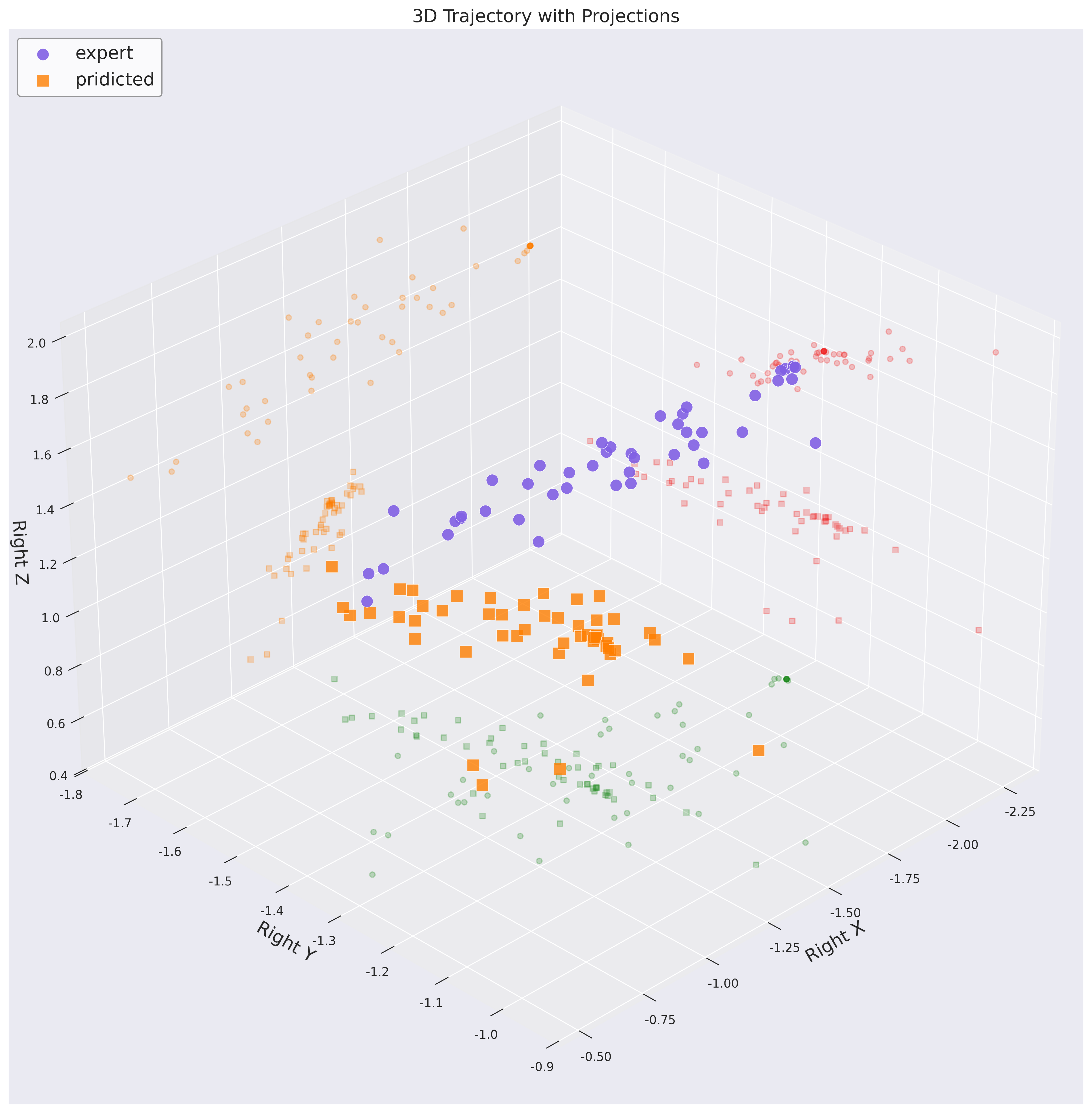}
\caption{\revision{Layer 1: Predicted vs Expert trajectories on training data (In-distribution Action Reconstruction)}}
\end{figure}

\textbf{Methodology}: We randomly sample $N$ frames from the training dataset. For each frame, we input the original observation to the policy and compare the predicted action with the ground truth expert action.

\textbf{Metrics and Interpretation}:

The Mean Absolute Error (MAE) measures average prediction error:
\begin{equation}
    \text{MAE} = \frac{1}{N} \sum_{i=1}^{N} |\hat{a}_i - a_i|
\end{equation}
MAE provides an intuitive measure of prediction accuracy in the same units as the action space. A MAE below 0.05 typically indicates good reconstruction quality for bimanual manipulation tasks.

The Mean Squared Error (MSE) penalizes large errors more heavily:
\begin{equation}
    \text{MSE} = \frac{1}{N} \sum_{i=1}^{N} (\hat{a}_i - a_i)^2
\end{equation}
MSE is useful for detecting occasional large errors that might not significantly affect MAE.

The Expert Similarity metric normalizes MAE by the variance of expert actions:
\begin{equation}
    \text{Expert Similarity} = 1 - \frac{\text{MAE}}{\sigma(a) + \epsilon}
\end{equation}
This provides a scale-invariant measure where 1.0 indicates perfect reconstruction and 0.0 indicates the policy prediction has no correlation with expert actions.

\textbf{Per-Dimension Analysis}: Beyond aggregate metrics, we compute per-dimension MAE to identify which action components are problematic:
\begin{equation}
    \text{MAE}_d = \frac{1}{N} \sum_{i=1}^{N} |\hat{a}_{i,d} - a_{i,d}|
\end{equation}
This is particularly useful for diagnosing specific issues, such as poorly calibrated gripper actions or incorrect rotation representations.

\textbf{Pass Criteria}: We empirically determine that $\text{MAE} < 0.05$ serves as a reasonable threshold for bimanual manipulation. However, this should be adapted based on the specific task and action space.

\section{Layer 2: Single-Step Closed-Loop Validation}

Beyond reconstruction accuracy, Layer 2 verifies that policy outputs are physically reasonable and smooth. This is particularly important because even accurate predictions can exhibit problematic patterns that would cause failures in real-time execution.

\textbf{Basic Statistics}: We first verify that output distributions match expected ranges:
\begin{align}
    \mu &= \frac{1}{N} \sum_{i=1}^{N} a_i \quad \text{(mean)} \\
    \sigma &= \sqrt{\frac{1}{N} \sum_{i=1}^{N} (a_i - \mu)^2} \quad \text{(standard deviation)} \\
    \text{range} &= \max(a) - \min(a)
\end{align}
If the predicted action distribution significantly differs from training data, it may indicate normalization issues or training instability.

\textbf{Action Smoothness}: Smooth motions are essential for stable robotic manipulation. We compute the action difference between consecutive timesteps:
\begin{equation}
    \Delta a_t = |a_{t+1} - a_t|
\end{equation}
Large action differences can cause mechanical stress and unstable control.

\textbf{Jerk Analysis}: Beyond action differences, we analyze jerk (the third derivative of position), which measures smoothness at the acceleration level:
\begin{equation}
    \text{Jerk} = \left|\frac{d^3 a}{dt^3}\right| = |a_{t+2} - 3a_{t+1} + 3a_t - a_{t-1}|
\end{equation}
High jerk values indicate jerky motions that can cause oscillation or instability in feedback control.

\textbf{Smoothness Score}: We compute a normalized smoothness score:
\begin{equation}
    \text{Smoothness Score} = \frac{1}{1 + \text{Jerk}}
\end{equation}
This provides a scale-invariant measure where higher values indicate smoother motions.

\textbf{Physical Validity}: We verify that predicted actions fall within physically reasonable bounds:
\begin{align}
    \text{position limits} &: a_{\text{min}} \leq a \leq a_{\text{max}} \\
    \text{velocity limits} &: |\Delta a| \leq v_{\text{max}} \\
    \text{acceleration limits} &: |\Delta^2 a| \leq \alpha_{\text{max}}
\end{align}

\textbf{Action Energy Statistics}: For manipulation tasks, action energy provides insight into required actuator effort:
\begin{equation}
    E = \sum_{d=1}^{D} a_d^2 = \|a\|_2^2
\end{equation}

\section{Layer 3: Short-Horizon Rollout Validation}

While Layers 1-2 verify per-timestep behavior, Layer 3 tests temporal consistency - whether errors accumulate over time. This simulates closed-loop execution where each prediction affects subsequent observations.

\textbf{Method}: Starting from an initial state, we run $K$ steps of model prediction, using ground-truth observations at each step. This isolates policy behavior from observation noise, focusing purely on action prediction consistency.

\textbf{Error Trajectory}:
\begin{align}
    E(t) &= \text{mean}(|\hat{a}_t - a_t|) \\
    \text{Final Error} &= E(K) \quad \text{(error at end of rollout)} \\
    \text{Error Growth} &= E(K) - E(0)
\end{align}

\textbf{Interpretation}: Negative error growth indicates the policy may be converging toward correct behavior, while positive growth indicates accumulating errors. Large error growth suggests the policy lacks consistent temporal understanding.

\textbf{Pass Criteria}: $\text{Error Growth} < 0.1$ indicates stable short-horizon behavior. However, exact thresholds depend on task horizon and action space.

\textbf{Robustness Test}: This layer can be extended by adding noise to observations during rollout to test robustness to sensor noise.

\section{Layer 4: Consistency Evaluation}

For stochastic policies or policies with stochastic elements (such as VAE-based policy), output consistency is crucial for reproducible behavior.

\textbf{Method}: We repeat inference $K$ times with identical observations and measure output variance.

\textbf{Variance Metrics}:
\begin{align}
    \text{Var}(a) &= \frac{1}{K} \sum_{k=1}^{K} (a_k - \bar{a})^2 \\
    \text{Mean Variance} &= \frac{1}{D} \sum_{d=1}^{D} \text{Var}(a_d)
\end{align}

\textbf{Consistency Score}: We compute a normalized consistency measure:
\begin{equation}
    \text{Consistency Score} = 1 - \min(\text{Mean Variance}, 1.0)
\end{equation}

\textbf{Noise Dependence Classification}:

\begin{table}[h]
\centering
\begin{tabular}{ccc}
\hline
Level & Variance Range & Interpretation \\
\hline
Very Low & $< 0.001$ & Essentially deterministic, suitable for precision tasks \\
Low & $0.001 - 0.01$ & Minor variations, acceptable for most manipulation tasks \\
Medium & $0.01 - 0.05$ & Noticeable variability, may affect precision tasks \\
High & $0.05 - 0.1$ & Significant variations, requires evaluation for specific task \\
Very High & $> 0.1$ & Unstable output, problematic for most tasks \\
\hline
\end{tabular}
\end{table}

For deterministic policies (such as ACT without VAE), variance should be essentially zero. Non-zero variance may indicate numerical instability or GPU precision issues.

\end{appendices}

\end{document}